\documentclass[preprint,12pt]{elsarticle}

\usepackage{amssymb}
\usepackage{amsmath}
\usepackage{textcomp}
\usepackage{multirow}
\usepackage{booktabs}
\newcommand{\circledletter}[1]{\textcircled{\scriptsize #1}}

\journal{Nuclear Physics B}

\begin{document}

\begin{frontmatter}



\title{Learning to Reconstruct Temperature Field from Sparse Observations with Implicit Physics Priors}


\author[label1,label2,label3]{Shihang Li} 
\ead{lishh_07@sjtu.edu.cn}
\author[label2,label3]{Zhiqiang Gong}
\ead{gongzhiqiang13@nudt.edu.cn}
\author[label2,label3]{Weien Zhou}
\ead{weienzhou@outlook.com}
\author[label1]{Yue Gao}
\ead{yuegao@sjtu.edu.cn}
\author[label2,label3]{Wen Yao}
\ead{wendy0782@126.com}
\affiliation[label1]{organization={MoE Key Lab of Artificial Intelligence, AI Institute, Shanghai Jiao Tong University},
            city={Shanghai 200240},
            country={China}
            }
\affiliation[label2]{
            organization={Defense Innovation Institute, Academy of Military Science},
            city={Beijing 10071},
            country={China}}

\affiliation[label3]{
            organization={Intelligent Game and Decision Laboratory},
            city={Beijing 10071},
            country={China}}

\begin{abstract}
Accurate reconstruction of temperature field of heat-source systems (TFR-HSS) is crucial for thermal monitoring and reliability assessment in engineering applications such as electronic devices and aerospace structures. However, the high cost of measurement acquisition and the substantial distributional shifts in temperature field across varying conditions present significant challenges for developing reconstruction models with robust generalization capabilities.
Existing DNNs-based methods typically formulate TFR-HSS as a one-to-one regression problem based solely on target sparse measurements, without effectively leveraging reference simulation data that implicitly encode thermal knowledge. To address this limitation, we propose IPTR, an implicit physics-guided temperature field reconstruction framework that introduces sparse monitoring–temperature field pair from reference simulations as priors to enrich physical understanding. To integrate both reference and target information, we design a dual physics embedding module consisting of two complementary branches: an implicit physics-guided branch employing cross-attention to distill latent physics from the reference data, and an auxiliary encoding branch based on Fourier layers to capture the spatial characteristics of the target observation. The fused representation is then decoded to reconstruct the full temperature field. Extensive experiments under single-condition, multi-condition, and few-shot settings demonstrate that IPTR consistently outperforms existing methods, achieving state-of-the-art reconstruction accuracy and strong generalization capability.
\end{abstract}




\end{frontmatter}



\section{Introduction}
\label{sec: intro}
Accurate perception of temperature distribution plays a pivotal role in ensuring the operational safety, efficiency, and long-term reliability of modern engineered systems~\cite{grujicic2005effect,zhao2023physics,yao2011review,he2021thermal,zhang2023multi}, particularly in electronic and aerospace where internal heat generation is substantial. With the ongoing trend toward higher integration and increased power density in electronic components~\cite{chen2016optimization}, the resulting thermal behavior becomes more intricate and difficult to predict, presenting significant challenges for thermal monitoring and control. 
In many practical scenarios-such as printed circuit boards~\cite{naphon2009numerical}, honeycomb panels in spacecraft~\cite{wang2019research},  satellite-mounted thermal components~\cite{chen2021deep,zhao2023physics}, and launch vehicle~\cite{benedikter2022convex}-only a limited number of temperature sensors can be deployed due to space, cost, or structural constraints. This sensor sparsity leads to a fundamental challenge: reconstructing the high fidelity temperature field of heat source systems from sparse point-wise measurements. This task, known as Temperature Field Reconstruction of Heat Source Systems(TFR-HSS), is inherently ill-posed~\cite{zhao2022state}, as the underdetermined nature of the observation makes it difficult to recover the full field through direct physical inversion.

Early efforts in TFR-HSS primarily adopted classical interpolation methods and traditional surrogate modeling techniques. Interpolation-based methods~\cite{capozzoli2010field}, such as linear, polynomial, and spline interpolation, estimate the temperature values in unsensed regions by fitting continuous functions over a discrete set of known sensor measurements. While these methods are computationally efficient and easy to implement, they neglect the intrinsic physical laws governing heat conduction, leading to poor accuracy and limited physical consistency. To address these limitations, traditional regression-based surrogate models have been introduced, including Kriging method~\cite{zhou2021insight}, support vector regression~\cite{razavi2019application}, radial basis function networks~\cite{ma2019neural}. These models can capture certain nonlinear relationships but still lack the representational capacity required for high-resolution or large-scale tasks. Shallow neural architectures, such as random vector functional-link (RVFL) networks~\cite{do2020integrative} and GMDH networks~\cite{menad2019modeling,ahmadi2019applying}, have also been explored to enhance flexibility, yet their limited depth and parameterization constrain their ability to serve as end-to-end approximators for complex thermal patterns.

In recent years, deep neural networks (DNNs) have emerged as powerful tools for physical field reconstruction~\cite{chen2023machine,peng2022deep,zhao2024hybrid,fukami2021global,santos2023development,zhang2024context,long2025fr}, benefiting from their strong expressive capacity and end-to-end learning capabilities. By learning complex mappings from sparse monitoring measurements to global temperature field, DNNs have shown superior performance in handling nonlinear dependencies and spatial heterogeneity. VoronoiCNN~\cite{fukami2021global} incorporate nearest-neighbor interpolation to reorganize irregularly distributed monitoring data into structured grids, thereby making it compatible with convolutional neural network models for spatial field recovery. VoronoiUnet~\cite{xia2023reliability} develops a Unet-based~\cite{ronneberger2015u} framework that incorporates Voronoi-based encoding and integrates uncertainty estimation, thereby improving the model's reliability in handling sparse and irregular measurements. 
In parallel, physics-informed neural networks (PINNs)~\cite{gong2023joint,liu2022temperature,chi2025heat} incorporate governing physics by embedding partial differential equation (PDE) residuals and boundary conditions into the loss function, thus enforcing physical consistency and improving generalization under data-scarce conditions. Recently, operator learning~\cite{lu2019deeponet,li2020fourier,li2020multipole,nelsen2021random,zhao2024recfno} has emerged as a powerful paradigm for modeling mappings between infinite-dimensional function spaces. RecFNO~\cite{zhao2024recfno} introduces three encoding schemes for sparse sensors and employs FNO~\cite{li2020fourier} to reconstruct the corresponding flow and heat fields from sparse observations. By learning the underlying functional relationships—often through frequency-domain representations—these approaches enable resolution-independent modeling of systems governed by PDEs.

Nonetheless, despite these advances, most existing DNNs-based methods formulate temperature field reconstruction as a one-to-one regression problem: given sparse monitoring data, the model directly maps the input to the corresponding full field. Such formulation tends to overlook a potentially rich source of information—reference simulation data collected under similar thermal conditions—which may implicitly encode the spatial and physical regularities of the system. Effectively leveraging these priors can provide additional guidance and improve reconstruction accuracy, particularly under sparse observation scenarios where the inverse problem is inherently ill-posed.

To this end, we propose a novel implicit physics-guided temperature field reconstruction (IPTR) framework that conditions the reconstruction process on a reference pair comprising sparse measurements and its corresponding full-field distribution. By leveraging this prior information, the model is encouraged to capture transferable physical patterns from thermally similar states, enriching the reconstruction process with inductive physical priors. This approach effectively bridges the gap between purely data-driven and physics-informed paradigms by embedding learned physical consistency through structural similarity. Specifically, our framework consists of three key components. First, the Sparse Field Encoding module transforms discrete sensor measurements into a continuous spatial representation through Voronoi-based interpolation, preserving the geometric structure of the sensing layout and ensuring spatial alignment. Next, the Dual Physics Embedding Module, which forms the core of our architecture, processes both the reference and the target inputs through two parallel branches: an Implicit Physics-Guided Branch that utilizes cross-attention to extract implicit physical knowledge from the reference pair, and an Auxiliary Encoding Branch that employs a Fourier-based encoder to capture the spatial characteristics of the target input. This dual-branch design enables the model to integrate global physical priors with local observation patterns into a unified representation. Finally, the fused features are passed to the Field Decoding Module, which reconstructs the full-resolution temperature field over the 2D spatial domain.
This modular architecture enables IPTR to incorporate implicit physical priors from reference data, leading to improved reconstruction accuracy and enhanced generalization in multi-condition and few-shot transfer scenarios. Our main contributions can be summarized as follows:
\begin{itemize}
    \item We propose IPTR, a novel framework that leverages reference simulation pairs—comprising sparse measurements and their corresponding full field—to embed implicit physical priors into the reconstruction process. By conditioning on such reference information, the model captures transferable thermal patterns, which significantly improve reconstruction fidelity in extreme regions and enhance generalization under limited data scenarios.
    \item To simultaneously leverage reference data and target monitoring data, we design a Dual Physics Embedding Module which consists of two branches: an implicit physics-guided branch using cross-attention to extract implicit physical priors from reference data, and an auxiliary encoding branch based on Fourier layers to preserve target field characteristics.
    \item Extensive experiments across single-condition, multi-condition, and few-shot settings demonstrate that IPTR achieves state-of-the-art accuracy and strong generalization capability.
\end{itemize}

This paper is structured as follows: Section~\ref{sec: relate} introduces the temperature field reconstruction of heat-source systems and our motivation. Section~\ref{sec: method} presents the proposed IPTR framework in detail, including the sparse field encoding, dual physics embedding module and temperature field decoding. Section~\ref{sec: exp} demonstrates the effectiveness of our method through experiments under single-condition, multi-condition, and few-shot settings. Finally, Section~\ref{sec: conclusion} concludes the paper with a summary of key findings.

\section{Temperature Field Reconstruction of Heat-Source Systems(TFR-HSS)}
\label{sec: relate}
\subsection{Mathematical Definition}
This work focuses on the heat-source systems comprising multiple internal heat transfer predominantly occurs via conduction within a two-dimensional spatial domain. To facilitate analysis while preserving generality, the system is modeled as a planar region containing localized heat source, each corresponding to a discrete component, such as an electronic component~\cite{chen2023machine,gong2023joint}. Within this framework, the problem of Temperature Field Reconstruction for Heat Source Systems (TFR-HSS) involves inferring the full temperature distribution across the domain based solely on a limited number of temperature measurements obtained from strategically deployed sensors. 
\begin{figure}[t]
\centering
\includegraphics[width=0.55\linewidth]{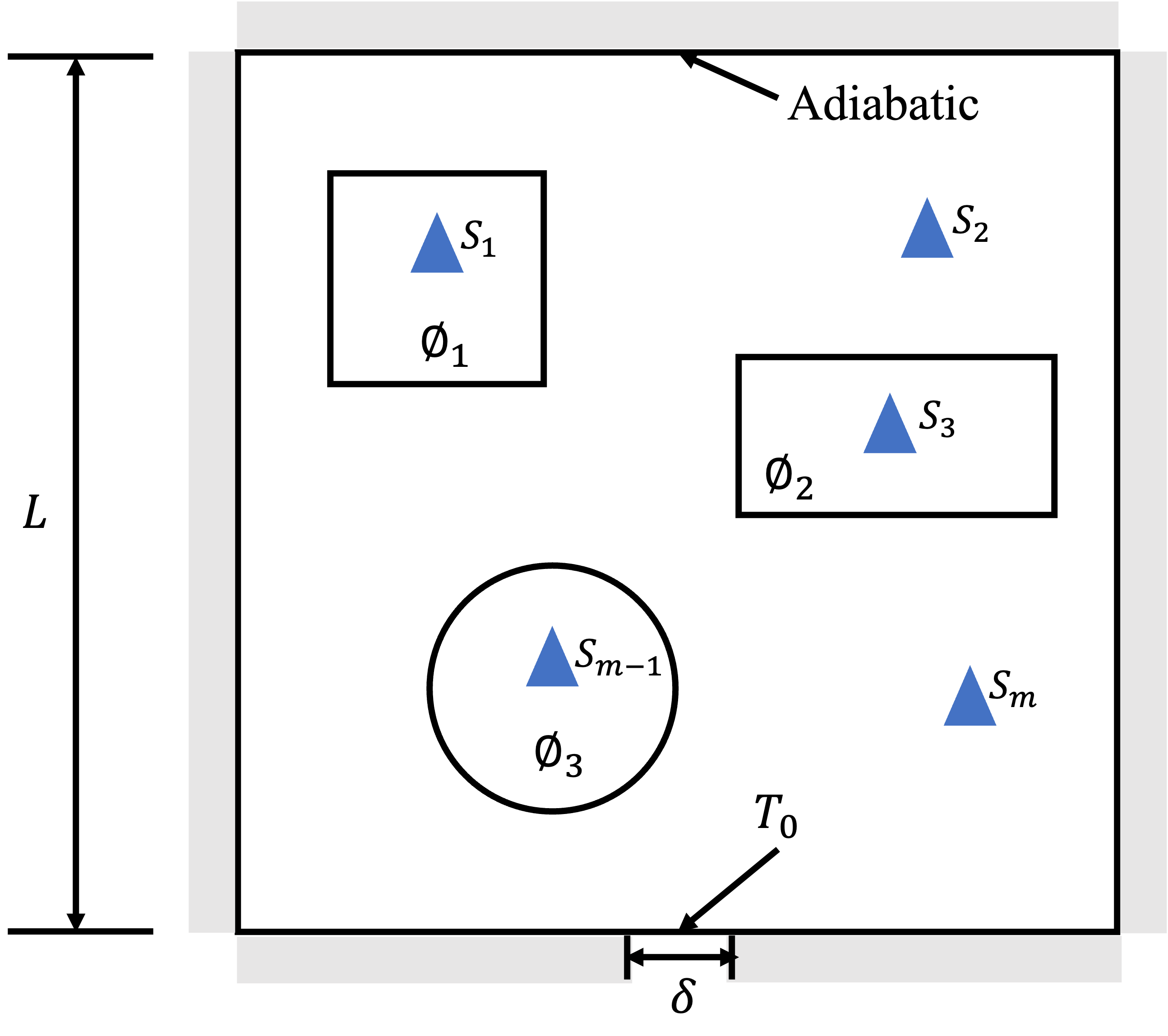}
\caption{An illustration of the region $\Omega$ of the HSS subject to Dirichlet boundary condition.}\label{hss}
\end{figure}
Let $\Omega \subset \mathbb{R}^2$ denote a bounded spatial domain representing the layout of the HSS. Within this domain, there exists $\Lambda$ distinct heat sources, each denoted by an intensity function $\phi_i(x,y)$, which characterizes the internal heat generation of the $i$-th source, as shown in Figure \ref{hss}. The temperature field is represented by a scalar function $T(x,y)$, and the spatially varying thermal conductivity is denoted by $\lambda$. Under steady-state conditions, the temperature distribution satisfies the stationary heat conduction equation with internal heat sources, which takes the form of Laplace equation:
\begin{equation}
\frac{\partial}{\partial x}\left(\lambda \frac{\partial T}{\partial x}\right) 
+ \frac{\partial}{\partial y}\left(\lambda \frac{\partial T}{\partial y}\right) 
+ \sum_{i=1}^{\Lambda} \phi_i(x, y) = 0, 
\quad (x, y) \in \Omega
\label{pde}
\end{equation}
subject to appropriate boundary conditions, which include the Dirichlet boundary condition (Dirichlet B.C.), the Neumann boundary condition (Neumann B.C.), and the Robin boundary condition (Robin B.C.). These can be expressed as:
\begin{equation}
T=T_0 \;(\text{Dirichlet}),\text{or}\; \lambda \frac{\partial T}{\partial n}=0 \;(\text{Neumann}), \text{or}\; \lambda \frac{\partial T}{\partial n}=h(T-T_0) \;(\text{Robin}),
\label{condition}
\end{equation}
where $n$ is the outward normal on the boundary, and $h$ is the convective heat transfer coefficient.
The objective of TFR-HSS task is to estimate the full temperature distribution $T(x,y)$ over $\Omega$ using only sparse monitoring measurements collected from $m$ sensors deployed at locations $\{(x_{s_i},y_{s_i})\}^m_{i=1}$. The corresponding observed temperature values are denoted as $\{f(x_{s_i},y_{s_i})\}^m_{i=1}$. This leads to the following inverse problem formulation:
\begin{equation}
T^* = \mathop{\text{argmin}}\limits_{T} \sum_{i=1}^{m} \left| T\left(x_{s_i}, y_{s_i} \middle| \phi_1, \dots, \phi_\Lambda \right) - f\left(x_{s_i}, y_{s_i}\right) \right|
\end{equation}
subject to the Equation (\ref{pde}) and (\ref{condition}). This problem is inherently ill-posed, owing to the limited number of observations and the high dimensionality of the target temperature field. To tackle this challenge, we aim to construct a data-driven regression model that learns the underlying mapping from sparse monitoring measurements to the full temperature field. In the following, we provide a formal problem formulation and outline the motivation of our approach.
\subsection{Problem Formulation and Motivation}
\begin{figure}[t]
\centering
\includegraphics[width=1\linewidth]{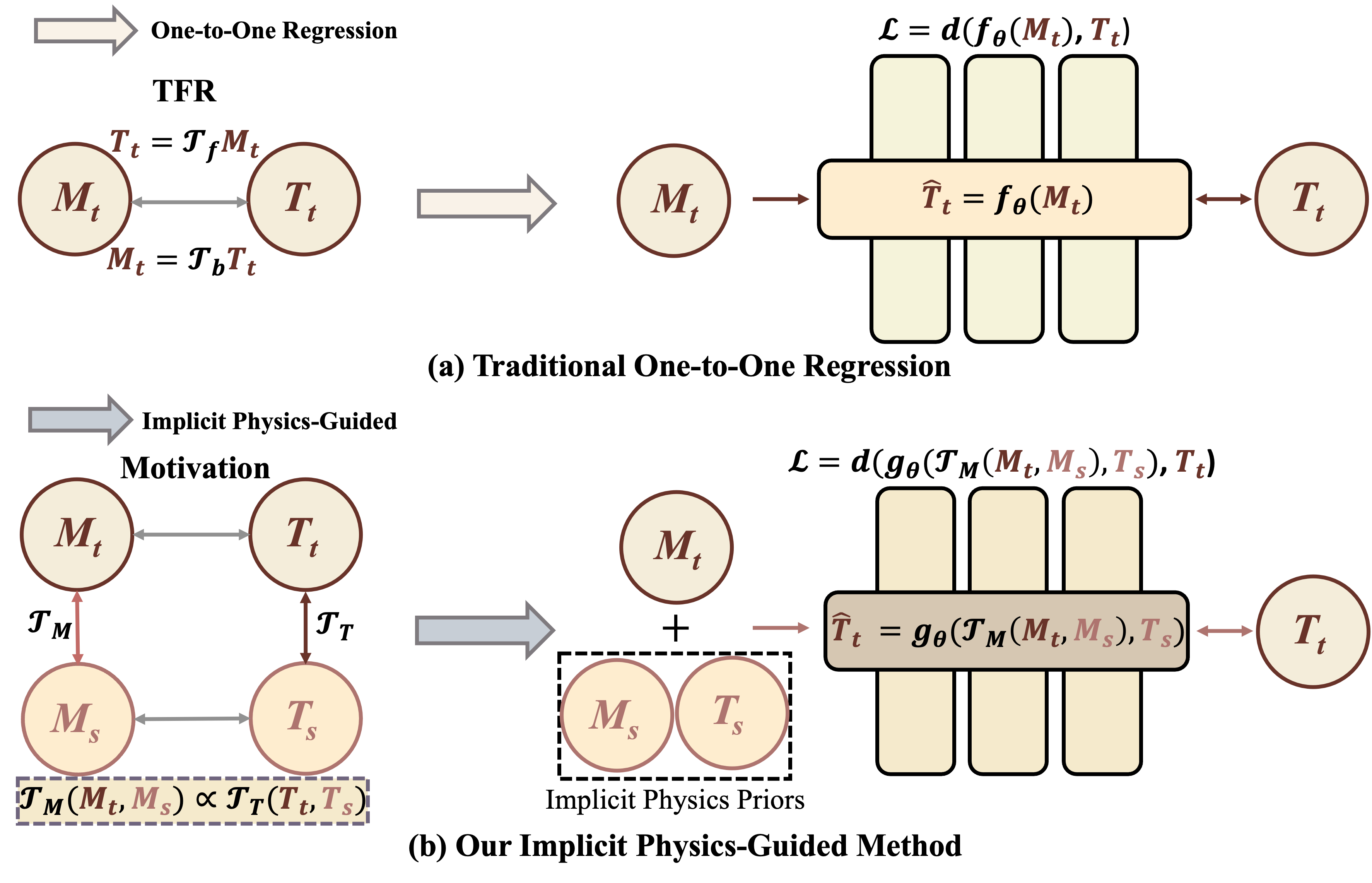}
\caption{Overview of our reconstruction paradigm. (a) Existing methods directly regress the temperature field $T_t$ from sparse measurements $M_t$, ignoring auxiliary information. (b) Our approach leverages a reference pair $(M_s, T_s)$ from similar thermal conditions to provide implicit physics guidance. By modeling the relation between $(M_t, M_s)$ and $(T_t, T_s)$, the network learns to reconstruct $T_t$ via the mapping $g_\theta(M_t, M_s, T_s) \rightarrow \hat{T_t}$.}\label{motivation}
\end{figure}

The TFR-HSS involves predicting the full-field temperature distribution $T_t$ from sparse monitoring measurements $M_t$. Notably, $M_t$ can also be regarded as a sampled subset of $T_t$, i.e., $M_t=\mathcal{T}_b T_t$. Existing approaches typically treat this process as a one-to-one regression, learning a direct mapping from $M_t$ to $T_t$ via a surrogate model $f_\theta$, as illustrated in Figure \ref{motivation}(a). While effective in controlled settings, such formulations ignore valuable prior knowledge available in practical scenarios, such as prior simulations data and their associated thermal conditions.
To address this limitation, we propose a new paradigm that introduces an additional reference pair—sparse monitoring measurements $M_s$ and its corresponding full-field temperature $T_s$ under a similar thermal condition—to implicitly embed physical guidance into the reconstruction process. This strategy allows the model to exploit thermally relevant reference information under similar condition, providing a form of implicit physics guidance that supplements the sparse input $M_t$.

Our motivation, shown in Figure \ref{motivation}(b), is based on the assumption that a transformation between $M_t$ and $M_s$, denoted as $\mathcal{T}_M(M_t,M_s)$, should correspond to a similar transformation $\mathcal{T}_T(T_t,T_s)$ between the associated temperature fields. \textbf{This implicit proportional relationship enables us to reformulate the reconstruction task as a function of both the target and the reference data, rather than treating them in isolation.}

To this end,  we define a neural network $g_\theta$ that models the mapping:
\begin{equation}
g_\theta : (M_t,M_s,T_s) \rightarrow \hat{T_t},
\end{equation}
where $g_\theta$ is parameterized by $\theta$, and $\hat{T_t}$ is the predicted temperature field. The model is trained to minimize the discrepancy between $\hat{T_t}$ and the ground truth $T_t$.
\section{Methodology}
\label{sec: method}
\begin{figure}[t]
\includegraphics[width=1\linewidth]{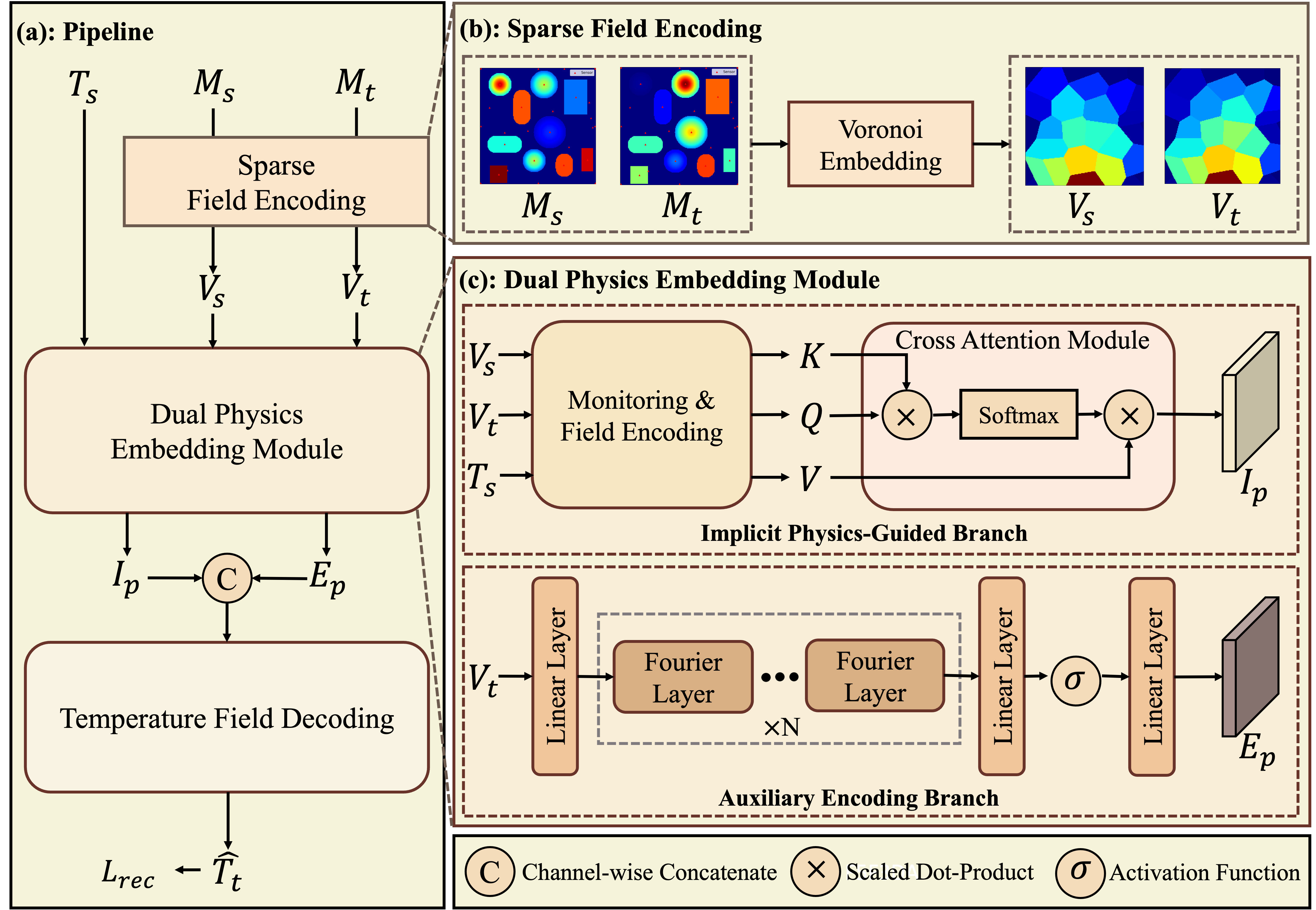}
\caption{Architecture of the proposed IPTR framework. (a) shows the overall pipeline for temperature field reconstruction. Given a reference pair consisting of sparse monitoring data $M_s$, and corresponding temperature field $T_s$, along with target sparse monitoring data $M_t$, we first obtain interpolated field $V_s$, $V_t$ via Voronoi-based encoding (b). These are passed to the Dual Physics Embedding Module (c), where the Implicit Physics-Guided Branch performs cross-attention among $V_s$, $T_s$ and $V_t$ to extract $I_p$, and the Auxiliary Encoding Branch encodes $V_t$ into $E_p$. The fused representation $I_p$\circledletter{C}$E_p$ is decoded to reconstruct the full temperature field $\hat{T_t}$, supervised by the ground-truth $T_t$.}
\label{fig: framework}
\end{figure}


In this section, we begin by presenting an overview of the proposed framework, which takes as input the target sparse monitoring data together with a reference pair of sparse monitoring–temperature field as priors. These reference data serve as implicit physical priors that guide the reconstruction process and enhance the model’s generalization capability (Section~\ref{sec: overview}).
Section~\ref{sec: sfe} introduces the Sparse Field Encoding module, where the sparse monitoring measurements are transformed into structured spatial representations using Voronoi-based interpolation.
The core component of our approach is the Dual Physics Embedding Module (Section~\ref{sec: dpem}), which effectively integrates the implicit physical knowledge encoded in the reference data with the target observations, through two complementary branches: the Implicit Physics-Guided Branch and the Auxiliary Encoding Branch.
Finally, the fused representation is passed to the Temperature Field Decoder, which reconstructs the full temperature field (Section~\ref{sec: tfd}).
\subsection{Overview of Our Framework}
\label{sec: overview}
As illustrated in Figure~\ref{fig: framework}, the proposed IPTR framework aims to reconstruct the full temperature field from sparse monitoring measurements by leveraging reference data from a similar thermal condition. The overall architecture is composed of three key components: the Sparse Field Encoding module, the Dual Physics Embedding module, and the Temperature Field Decoder. Given the target sparse monitoring temperature value and a selected pair of reference sparse monitoring-temperature field data, the framework first performs Voronoi-based interpolation to transform both the target and reference monitoring data into spatially structured fields. These interpolated fields are the passed into the Dual Physics Embedding module, which constitutes the core of our design. It consists of two parallel branches: the Implicit Physics-Guided Branch, which encodes reference information via a cross-attention mechanism to provide implicit physical priors; and the Auxiliary Encoding Branch, which encodes the target monitoring data using a Fourier-based encoder to enrich the learned representation. The features from both branches are fused and subsequently fed into the Temperature Field Decoder, which maps the unified representation to a high fidelity temperature field reconstruction. By jointly utilizing reference physical patterns and target observations in a unified learning framework, IPTR enhances reconstruction accuracy and spatial coherence by implicitly leveraging physical priors, and enables effective adaptation to unseen scenarios through few-shot fine-tuning.

\subsection{Sparse Field Encoding}
\label{sec: sfe}
The sparse and irregular distribution of sensors across the spatial domain introduces challenges for structured modeling of physical fields. Directly using raw monitoring data as model input often leads to poor spatial correlation modeling, as the lack of consistent spatial structure hinders effective information propagation and attention computation. 

To address this, we employ a Voronoi-based interpolation strategy ~\cite{fukami2021global,zhao2024recfno} to transform the sparse monitoring measurements into structured grid-aligned representations. Let the target sparse monitoring measurements set be denoted as $M_t=\{(p_i^t,m_i^t)\}^{N_t}_{i=1}$, where $p_i^t \in \Omega \subset \mathbb{R}^2$ is the spatial location of the $i$-th sensor and $m_i^t\in \mathbb{R}$ denotes its corresponding temperature value. Here, $\Omega$ is discretized into a uniform 2D grid $\mathcal{G}=\{(x_i,y_i)|i=1,...,H;j=1,...,W\}$, which defines the spatial resolution of the reconstructed field.  

To propagate sparse monitoring measurements to the full grid, we partition $\Omega$ using a Voronoi tessellation induced by the sensor locations $\{(p_i^t)\}^{N_t}_{i=1}$. Each grid point $(x_i,y_i)\in \mathcal{G}$ is assigned the temperature value of its nearest sensor in Euclidean distance, i.e.,
\begin{equation}
V_t(i,j) = m_k \quad \text{where} \quad 
k = \arg\min_{1 \leq \ell \leq N_t} \left\| (x_i, y_j) - \mathbf{p}_\ell^t \right\|_2
\end{equation}

\begin{figure}[t]
\includegraphics[width=1\linewidth]{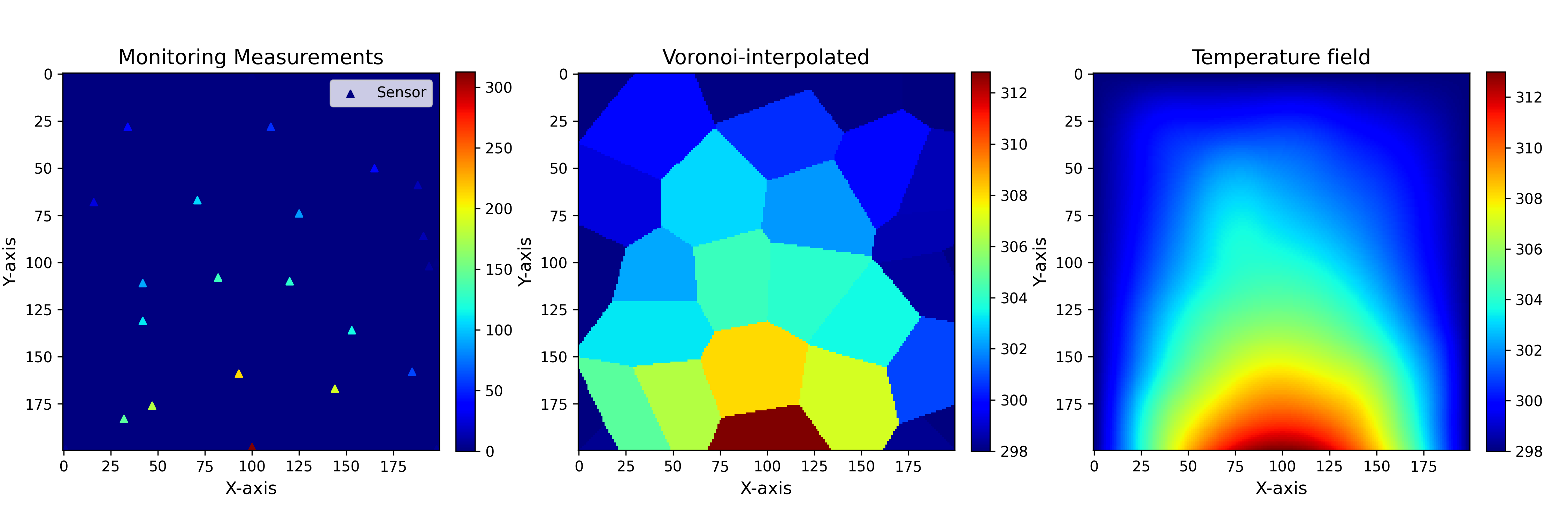}
\caption{Voronoi-based pseudo-field generation.}
\label{fig: vor}
\end{figure}
This process yields a fully populated 2D pseudo-field $V_t \in \mathbb{R}^{H\times W}$, referred to as the Voronoi-interpolated temperature map. A similar procedure is applied to the reference measurements $M_s$, resulting in $V_s \in \mathbb{R}^{H\times W}$. 
Figure~\ref{fig: vor} illustrates this process, where sparse and irregular measurements are first assigned to a uniform spatial grid via Voronoi tessellation, producing a structured pseudo-field that approximates the underlying temperature distribution. This transformation improves spatial continuity while preserving the sensor layout, enabling grid-aligned model inputs.
The structured representations $(V_t, V_s)$ facilitate subsequent cross-attention computation by aligning both the target and reference monitoring measurements to a common spatial grid. Compared to unordered sensor inputs, Voronoi interpolation enhances spatial consistency and interpretability, ultimately improving the model’s ability to infer underlying physical patterns under sparse sensing conditions, as discussed in Section~\ref{sec: dpem}.
\subsection{Dual Physics Embedding Module}
\label{sec: dpem}
While the Voronoi-based interpolation provides a structured representation of the sparse monitoring data, it alone lacks the ability of capture deeper physical dependencies and correlations necessary for accurate reconstruction. To address this, we propose the Dual Physics Embedding Module, which forms the core of our framework. Its key motivation is to leverage reference sparse monitoring-temperature field as implicit physical priors, enabling the model to reason beyond the target sparse monitoring measurements and generalize better in regions with complex thermal variations.

As illustrated in Figure~\ref{fig: framework}(c), this module consists of two complementary branches designed to encode both the target interpolated field and reference information. The resulting embeddings are fused to form a unified representation for temperature field reconstruction. In the following, we detail the designs of the Implicit Physics-Guided Branch and the Auxiliary Encoding Branch.

\subsubsection{Implicit Physics-Guided Branch}
The goal of this branch is to inject implicit physical priors derived from a reference pair into the reconstruction process. Rather than explicitly modeling physical laws, we encode these priors by utilizing a reference sparse monitoring-tempearature field pair ($M_s$, $T_s$), which captures thermal behaviors under similar thermal conditions.

To model the correlation between the target sparse measurements $M_t$ and the reference measurements $M_s$—as conceptualized in Figure \ref{motivation}(b)—we employ a cross-attention mechanism. Specifically, we assume that the similarity between $M_t$ and $M_s$ can guide the transfer of information from $T_s$ to assist in reconstructing the target temperature field $T_t$. In practice, this is implemented via a single scaled dot-product attention~\cite{vaswani2017attention} operating on the target interpolated field $V_t$ as the query, and the reference interpolated field $V_s$ and full field $T_s$ as the key and value, respectively. The attention operation is defined as:
\begin{equation}
Attention(Q,K,V)=Softmax(\frac{Q\times K^T}{\sqrt{D}})\times V
\end{equation}
where $Q$, $K$ and $V$ denote the queries, keys and values, and $D$ is a scalar scaling factor equal to the feature dimension. By computing attention in this way, the model learns to dynamically associate target observations with relevant reference patterns in a data-driven yet physically grounded manner, effectively enhancing its reconstruction capability, particularly in spatial regions with complex thermal variations.

Considering computational resources and efficiency, we apply attention in the latent space, rather than the pixel space. To this end, we employ a monitoring encoder $\mathcal{E}_{m}$ to encode $V_s \in \mathbb{R}^{H\times W}$ and $V_t \in \mathbb{R}^{H\times W}$ into $\tilde{V}_s \in \mathbb{R}^{C \times \frac{H}{4}\times \frac{W}{4}}$ and $\tilde{V}_t \in \mathbb{R}^{C \times \frac{H}{4}\times \frac{W}{4}}$, and a temperature field encoder $\mathcal{E}_{tf}$ to encode $T_s \in \mathbb{R}^{H\times W}$ into $\tilde{T}_s \in \mathbb{R}^{C \times \frac{H}{4}\times \frac{W}{4}}$. Note that these encoders are all UNet-based~\cite{ronneberger2015u} encoders. 

We assign the latent of $\tilde{V}_t$ as the query because it represents the target to be reconstructed-it is where the model should attend to relevant information. The latent of reference monitoring data $\tilde{V}_s$ is used as the key, providing a spatial anchor from reference measurements. Finally, the latent of reference temperature field $T_s$ serves as the value, containing rich physical structures that can be transferred to guide the reconstruction. Note that $Q$, $K$ and $V$ are all reshaped from $(C, \frac{H}{4}, \frac{W}{4})$ to $(\frac{H}{4} \times \frac{W}{4}, C)$, we also add positional embeddings to $Q$ and $K$. Finally, the implicit-physics guided representation is formulated as:
\begin{equation}
I_p = Attention(Q,K,V)
\end{equation}
As illustrated in Figure~\ref{fig: framework}(c), this branch provides the core physical guidance for our model. The resulting embedding $I_p$ will be later fused with auxiliary features for downstream reconstruction.

We next introduce the Auxiliary Encoding Branch, which complements this component by providing structural enhancement.
\subsubsection{Auxiliary Encoding Branch}
\begin{figure}[htbp]
\includegraphics[width=1\linewidth]{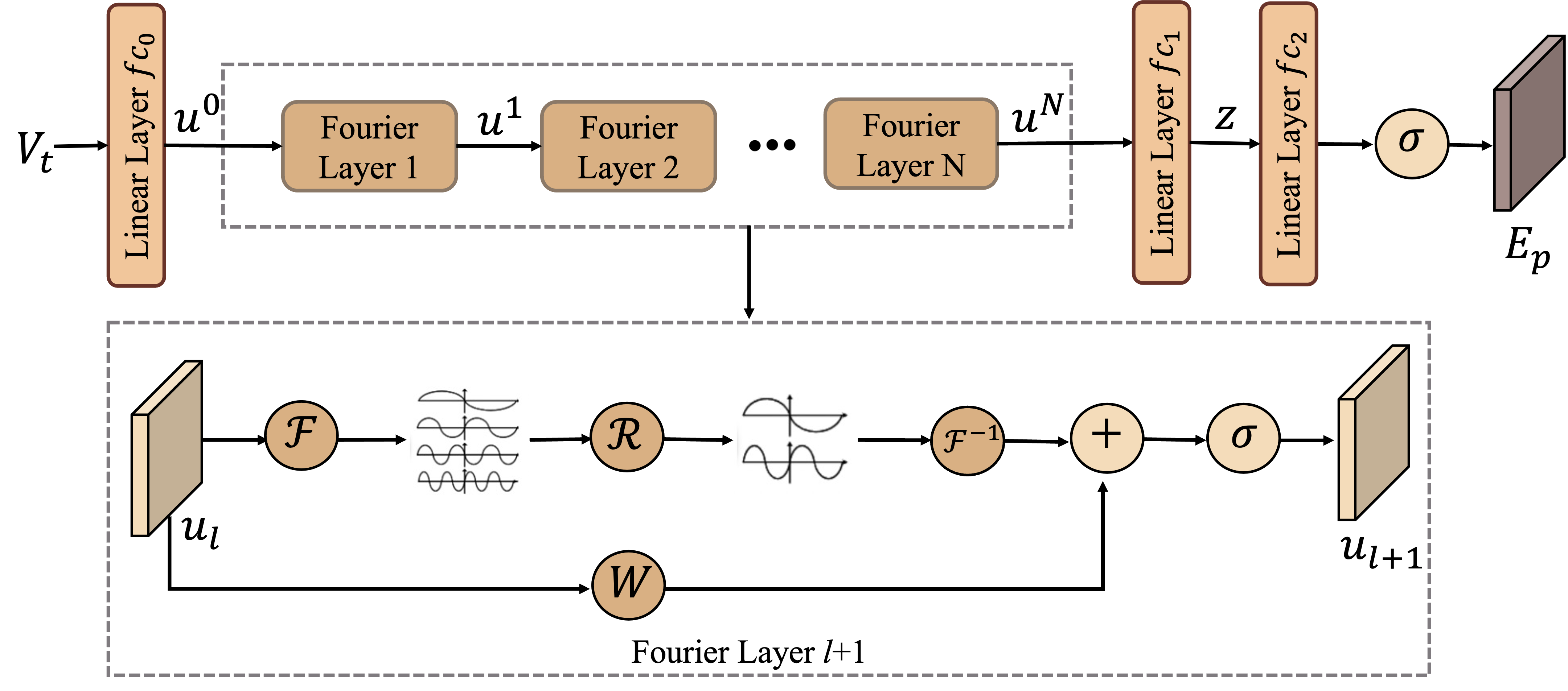}
\caption{Architecture of the Auxiliary Encoding Branch. The target sparse interpolated field $V_t$ undergoes a dimensional lifting, followed by spectral transformation through Fourier layers, and is subsequently projected to obtain auxiliary embeddings $E_p$ for enhancing the reconstruction process.}
\label{fig: fourier}
\end{figure}

To complement the implicit physics-guided reconstruction, we design an Auxiliary Encoding Branch that enhances the structural representation of the target input. Although not directly involved in physics prior extraction, this branch encodes the interpolated sparse field $V_t \in \mathbb{R}^{H \times W}$ into an auxiliary representation $E_p \in \mathbb{R}^{1 \times H \times W}$, which serves as a complementary source of spatial information. By capturing non-local correlations in the spectral domain, it enriches the feature space and promotes more robust and spatially consistent reconstruction.

As illustrated in Figure~\ref{fig: fourier}, this branch is based on a residual Fourier Neural Operator (FNO)~\cite{li2020fourier} structure, implemented using a stack of Fourier-based convolutional layers interleaved with residual connections and non-linear activations. The processing begins with a point-wise linear transformation which lift the input $V_t$ from a single-channel field into a high-dimensional latent space:
\begin{equation}
u^{(0)}=fc_0(V_t),\quad u^{(0)}\in \mathbb{R}^{C'\times H\times W}
\end{equation}
where $C'$ denotes the channel dimension of the lifted space. This lifting step enables the subsequent spectral operations to operate in a richer representational space.

Following the lifting, the latent feature map is successively updated through $N$ Fourier layers. In each layer, the update rule is defined as:
\begin{equation}
u^{(l+1)}(x)=\sigma(Wu^{(l)}(x)+\mathcal{K}(u^{(l)})(x))
\end{equation}
where $\sigma(\cdot)$ denotes the non-linear GELU activation function, $W$ is a learnable $1\times1$ convolution representing a local linear transformation, and $\mathcal{K}$ is a parameterized global operator defined in the frequency domain.

Specifically, the global operator $\mathcal{K}$ is formulated as a kernel integral operator:
\begin{equation}
\mathcal{K}(u)(x)=\mathcal{F}^{-1}(\mathcal{R} \cdot \mathcal{F}(u))(x),
\end{equation}
where $\mathcal{F}$ and $\mathcal{F}^{-1}$ denote the 2D Fourier transform and its inverse, and $\mathcal{R}\in \mathbb{C}^{F_1\times F_2\times C' \times C'}$ is a complex weight that operates as a spectral filter in the frequency domain, modulating selected low-frequency components $(F_1,F_2)$. For implementation efficiency, we truncate the spectrum and retain only the lowest $F_1$ and $F_2$ modes (i.e., low-pass filtering). The forward Fourier transform is performed as:
\begin{equation}
\hat{u}^{(l)}(k_1, k_2) = 
\sum_{x_1=0}^{H-1} \sum_{x_2=0}^{W-1} 
u^{(l)}(x_1, x_2) \cdot 
e^{-2\pi i \left( \frac{x_1 k_1}{H} + \frac{x_2 k_2}{W} \right)}
\end{equation}
while the inverse transform is:
\begin{equation}
u'^{(l)}(x_1, x_2) = 
\sum_{k_1=0}^{F_1-1} \sum_{k_2=0}^{F_2-1} 
(R \cdot \hat{u}^{(l)})(k_1, k_2) \cdot 
e^{2\pi i \left( \frac{x_1 k_1}{H} + \frac{x_2 k_2}{W} \right)}
\end{equation}
The multiplication in the frequency domain is defined element-wise for each mode $(k_1,k_2)$ and each output channel $i\in\{1,...,C'\}$:
\begin{equation}
(R \cdot \hat{u}^{(l)})_{k,i} = \sum_{j=1}^{C'} R_{k,i,j} \cdot \hat{u}^{(l)}_{k,j}
\end{equation}
where $k=(k_1,k_2)$.
After $N$ such Fourier layers (N=4), the resulting representation $u^{(N)}\in \mathbb{R}^{C'\times H\times W}$ is first mapped back to a lower dimensional channel space through two fully connected layers:
\begin{equation}
z=fc_1(u^{(N)}), \quad \tilde{E}_p = fc_2(\sigma(z)) 
\end{equation}
To ensure compatibility with the latent resolution of the implicit physics-guided branch (e.g., due to spatial downsampling by the encoder), we apply a spatial downsampling via max pooling:
\begin{equation}
E_p=\text{MaxPool}_{4\times 4}(\tilde{E}_p), \quad \text{where} \; \tilde{E}_p \in \mathbb{R}^{1\times H \times W}, \, E_p \in \mathbb{R}^{1\times \frac{H}{4} \times \frac{W}{4}}
\end{equation}
This projection ensures the dimensional alignment with downstream modules and allows the auxiliary output $E_p$ to be effectively fused with the implicit physics-guided features $I_p$ during decoding. The entire architecture thus learns to represent spatial patterns in both physical and frequency domains, offering complementary guidance to the reconstruction task.
\subsection{Temperature Field Decoding}
\label{sec: tfd}
The final stage of our framework aims to reconstruct the full temperature field from the learned physics-aware representations. To achieve this, we employ a decoding module that integrates both branches of the Dual Physics Embedding Module, effectively combining implicit physics prior knowledge and auxiliary Fourier-based features for high-fidelity reconstruction.

Specifically, we concatenate the implicit physics-guided features $I_p \in \mathbb{R}^{C\times \frac{H}{4} \times \frac{W}{4}}$ and the auxiliary encoded features $E_p \in \mathbb{R}^{1 \times \frac{H}{4} \times \frac{W}{4}}$ along the channel dimension, resulting a unified representation:
\begin{equation}
F_p= \text{Concat}(I_p,E_p) \in \mathbb{R}^{(C''\times \frac{H}{4} \times \frac{W}{4})}
\end{equation}
To decode this fused representation into the final temperature reconstruction, we adopt a SPADE decoder~\cite{park2019semantic}, which has shown strong performance in structured generation tasks due to its ability to incorporate spatial conditioning in a learnable, dynamic manner.

Formally, let $\mathcal{D}_{\text{SPADE}}$ denote the SPADE decoder. The final reconstructed temperature field $\hat{T}_t$ is given by:
\begin{equation}
\hat{T}_t=\mathcal{D}_{\text{SPADE}}(F_p)
\end{equation}
where $\hat{T}_t \in \mathbb{R}^{1 \times H \times W}$ corresponds to the model's prediction of the dense thermal field under current observation. 

By integrating both $I_p$ and $E_p$ into a unified decoding process, the framework benefits from complementary information sources. The representation $I_p$ encodes implicit physical priors derived from reference pair through cross-attention, capturing spatial correlations that reflect underlying physical behaviors. In contrast, $E_p$ provides auxiliary spatial structure information extracted from the target interpolated sparse observations via a Fourier-based encoder, serving as a complementary cue to enhance local consistency and improve robustness under measurement sparsity. This design enables the decoding module to leverage both implicit physics knowledge and target structural features in a unified reconstruction pipeline.

\section{Experiments}
\label{sec: exp}
\subsection{Experiment setups}
\subsubsection{Experimental Objectives}

To comprehensively evaluate the performance and generalization capability of the proposed IPTR framework, we conduct experiments under two categories of settings:

\noindent \textbf{Benchmark Evaluation}: We first evaluate IPTR on the TFRD benchmark dataset~\cite{chen2023machine}, which consists of various heat source configurations. Under this setting, we conduct two experiments:
\begin{itemize}
    \item Single-condition reconstruction, where both training and testing are performed on the same heat configuration;
    \item Multi-condition reconstruction, where the model is trained on multiple heat configurations and tested on held-out ones within the same dataset.
\end{itemize}
\noindent \textbf{Generalization to New Scenario}: To evaluate the model's generalization ability to previously unseen configurations, we introduce a new heat configuration as a target domain. In this setting: we perform:
\begin{itemize}
    \item Sample-efficiency comparison, where we compare IPTR’s reconstruction performance with other models trained from scratch using varying numbers of samples;
    \item Few-shot transfer learning, where IPTR, pre-trained on the benchmark dataset, is fine-tuned using a limited number of labeled samples from the new scenario;
    \item Sensor sparsity analysis, where we evaluate the reconstruction performance of IPTR trained from scratch under different number of sensors in the new scenario.
\end{itemize}
\subsubsection{Datasets}
In this work, we employ two categories of datasets to evaluate the proposed method: the TFRD benchmark dataset and a newly constructed scenario dataset. Both datasets are generated based on the governing equation and boundary conditions defined in Equation (\ref{pde}) and (\ref{condition}), simulating steady-state heat conduction in a $0.1$m$\times0.1$m satellite panel. The temperature field is discretized into a 2D grid of resolution $200\times 200$, representing fine-grained thermal distributions across the panel.

To characterize typical thermal behaviors in HSS, two representative power distribution settings are widely used in engineering scenarios: uniformly distributed and ununiformly distributed heat sources. In the uniform setting, each heat-generating component maintains a constant power density across its surface, resulting in a spatially uniform thermal load. In contrast, the ununiform setting captures spatially varying heat generation, often modeled using a Gaussian distribution, where the power density peaks at the component center and decays towards the boundary. This can be formally expressed as:
\begin{equation}
\phi_i(x, y) = 
\begin{cases} 
    Q_i \exp\left( -\sigma \dfrac{(x - x_0)^2 + (y - y_0)^2}{r^2} \right), & (x, y) \in \Omega_i, \\
    0, & \text{otherwise}.
\end{cases}
\end{equation}
where $Q_i$ is the peak power, $(x_0,y_0)$ is the center of the $i$-th component, $r$ denotes the radius, and $\sigma$ is the deviation coefficient.

\begin{figure}[t]
\centering
\includegraphics[width=1\linewidth]{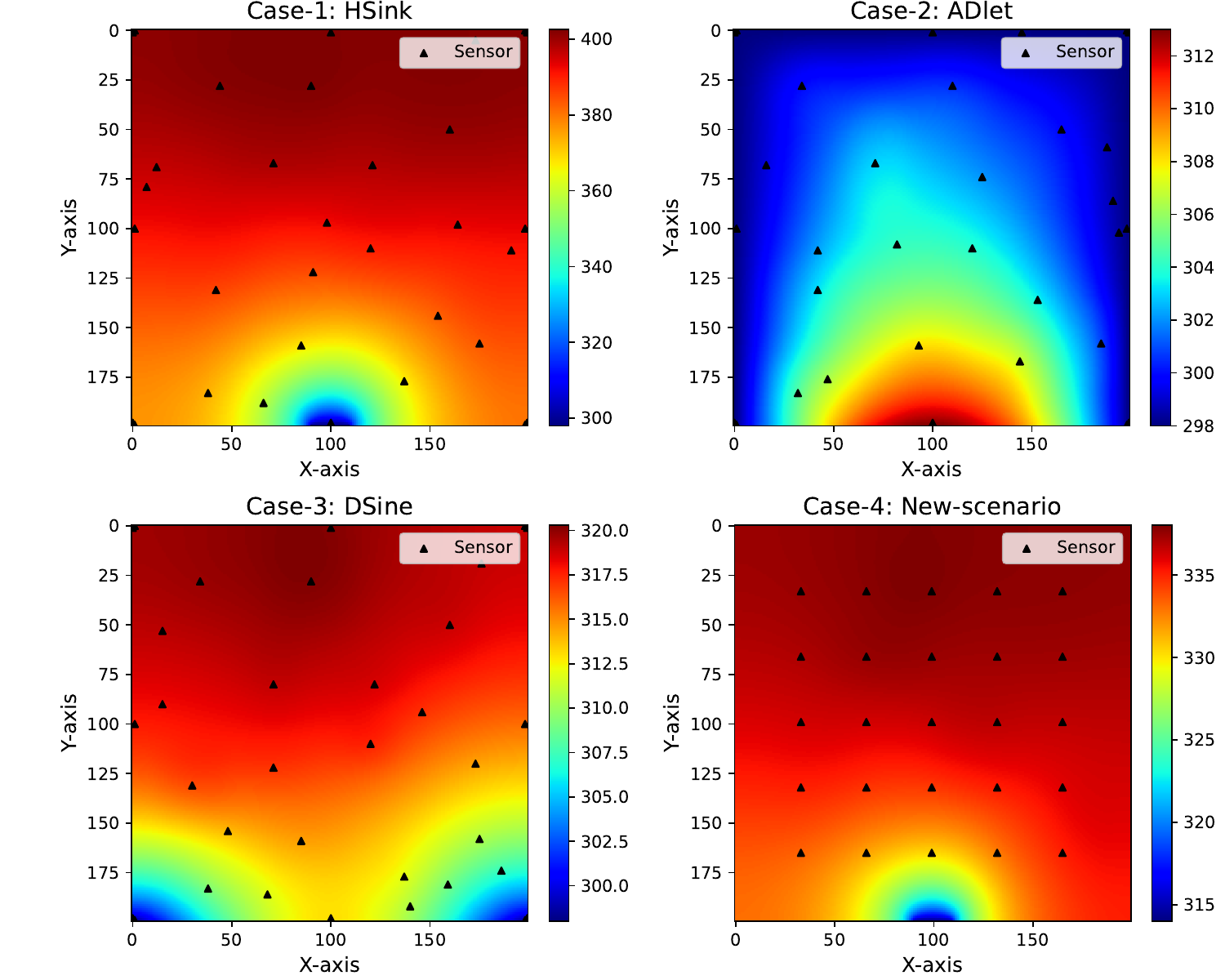}
\caption{Visualization of four datasets used in this paper}\label{tfr_example}
\end{figure}

The TFRD benchmark consists of three representative sub-tasks: HSink, ADlet and DSine, each defined by specific boundary conditions and power configurations: HSink applies a Dirichlet boundary with a fixed temperature of 298K on one side (heat sink) and Neumann (adiabatic) conditions on the remaining sides, with uniform power sources; ADlet imposes Dirichlet boundaries on all sides, where one side follows a sine-wave temperature distribution and others remain constant, using Gaussian-distributed heat sources; DSine adopts a single sine-wave Dirichlet boundary with the other three sides being adiabatic, also with Gaussian power sources. In addition, we construct a new scenario dataset with Gaussian heat sources and a sink-type Dirichlet boundary, the thermal conductivity is modeled as temperature-dependent: $\lambda=1+0.05\times (T-298)$. Furthermore, 25 temperature sensors are uniformly placed across the spatial domain to reflect sparse real-world monitoring. Visualizations of the four datasets are shown in Figure \ref{tfr_example}.
\subsubsection{Implementation and Evaluation Metrics}
\noindent \textbf{Implementation Details}: All models are implemented using the PyTorch framework and trained on a single NVIDIA 3090 GPU. In the benchmark evaluation experiments, the model is trained from scratch for 150 epochs with a batch size of 16. For few-shot adaptation in the new scenario, fine-tuning is conducted for 100 epochs. The batch size is set to 1 in few-shot fine-tuning experiments and remains 16 in all other settings. The Adam optimizer~\cite{kingma2014adam} is employed with an initial learning rate of 1.5e-4. A multi-step learning rate scheduler (MultiStepLR)~\cite{paszke2019pytorch} is used, where the learning rate is decayed by a factor of 0.1 at predefined epoch milestones. Specifically, the milestone is set to 100 for training from scratch, and 70 for fine-tuning.

\noindent \textbf{Evaluation Metrics}: To quantitatively evaluate the performance of our model in temperature field reconstruction, we adopt the following two commonly used metrics~\cite{chen2021deep}:
\begin{itemize}
    \item Mean Absolute Error (MAE). The MAE measures the average absolute deviation between the predicted temperature field $\hat{T}_t$ and the ground truth field $T_t$, and is defined as
    \begin{equation}
    \text{MAE}=\frac{||T_t-\hat{T}_t||}{|T_t|}
    \end{equation}
where $|T_t|$ denotes the total number of points in the field.

    \item Maximum Absolute error (Max-AE). The Max-AE reflects the worst-case deviation and is especially relevant for evaluating safety-critical systems involving extreme temperatures.
    \begin{equation}
    \text{MAE-AE}=||T_t-\hat{T}_t||_{\infty}
    \end{equation}
    
\end{itemize}
\subsection{Comparison with Other Methods}
To comprehensively evaluate the proposed IPTR framework, we conduct a series of experiments on both benchmark and newly constructed datasets. We compare our model with four representative baselines: Vor-UNet and Vor-FNO, which take Voronoi-interpolated fields as input; and Mask-UNet and Mask-FNO, which directly process raw sparse monitoring maps. UNet and FNO serve as the backbone architectures to isolate the impact of model architecture, while the two input settings distinguish the role of spatial interpolation. These baselines provide a fair and comprehensive comparison to our IPTR.
\subsubsection{Benchmark Evaluation}
\label{tfrd}
In the benchmark evaluation, we perform experiments under two settings: single-condition and multi-condition. We use the test set provided in the benchmark to evaluate model performance.
For the single-condition evaluation, we randomly sample 1,000 instances from each of the three TFRD sub-datasets (HSink, ADlet, and DSine). During training, we randomly form 1,000 reference–target pairs under the same thermal condition to supervise the reconstruction task. The specific sampling strategy is further discussed in Section \ref{aba}. For the multi-condition setting, we construct a combined training set by mixing the sampled instances from all three sub-datasets, allowing the model to learn under diverse boundary and source configurations. During testing, we randomly select one sample from the training set under the same thermal condition as the reference pair, and use all corresponding samples in the test set as target inputs for evaluation.

\noindent \textbf{Single-Condition Reconstruction}: As shown in Table \ref{tab:tfrd-single}, our method demonstrates consistently strong performance across all three cases based on quantitative metrics. In terms of MAE, the results of our method are 0.1154, 0.0149, and 0.0347 respectively—either achieving or closely approaching the best-performing models—while maintaining minimal variation across tasks, indicating robust generalization. More notably, our method achieves the lowest Max-AE values in all cases (3.2731, 0.1236, and 0.3207), outperforming the second-best methods by an average of 76.5\%, highlighting its superior ability to control local reconstruction errors. 
As the temperature span increases from 15.5K (ADlet) to 130.6K (HSink), most baseline methods show a significant rise in reconstruction error. For instance, Vor-UNet, as a representative baseline, sees its Max-AE grow from 1.1551 to 14.6286. In comparison, our method shows only a moderate increase—from 0.1236 to 3.2731—while still achieving the lowest error in both scenarios. This demonstrates that our model better preserves accuracy under more challenging temperature differences and maintains stable performance across varying conditions.
\begin{table*}[htbp]
\centering
\resizebox{1\linewidth}{!}{
\begin{tabular}{*{7}{c}} 
\toprule
\multirow{2}{*}{Method} & \multicolumn{2}{c}{\textbf{HSink}} & \multicolumn{2}{c}{\textbf{ADlet}} & \multicolumn{2}{c}{\textbf{DSine}} \\
\cmidrule(lr){2-3} \cmidrule(lr){4-5} \cmidrule(lr){6-7} 
& MAE$\downarrow$ & Max-AE$\downarrow$ & MAE$\downarrow$ & Max-AE$\downarrow$ & MAE$\downarrow$ & Max-AE$\downarrow$ \\
\midrule
Mask-UNet& 0.2715 & \underline{11.3475}
 & 0.0342 & \underline{0.6265} & 0.1148 & 2.7575  \\ 
Mask-FNO&1.4606 & 11.4818 & 0.4393 & 1.6913 & 2.1168 & 5.9253  \\ 
Vor-UNet& 7.7331 & 14.6286 & \textbf{0.0115} & 1.1551 & \underline{0.0350} & \underline{1.4806}  \\ 
Vor-FNO& \underline{0.2439} & 12.6760 & 0.0444 & 5.7005 & 0.1168 & 4.6461  \\ 
\midrule
Ours & \textbf{0.1154} &  \textbf{3.2731} & \underline{0.0149} & \textbf{0.1236} & \textbf{0.0347} & \textbf{0.3207} \\

\bottomrule 
\end{tabular}
}
\caption{Quantitative comparisons under single-condition reconstruction.}
\label{tab:tfrd-single}
\end{table*}

\begin{figure}[t]
\includegraphics[width=1\linewidth]{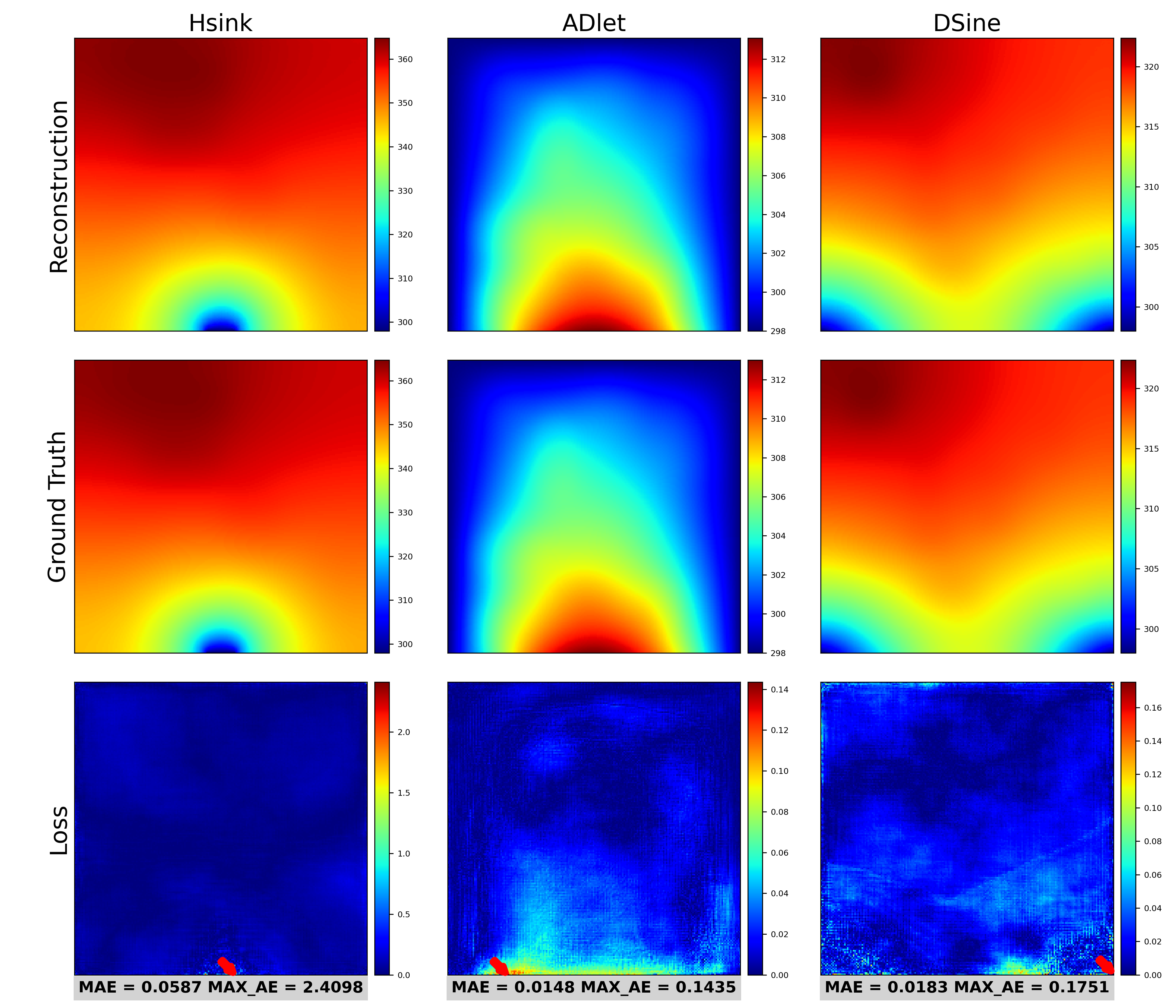}
\caption{Visualization of our result under singe-condition reconstruction.}
\label{fig: our_tfrd_single}
\end{figure}

\begin{figure}[htbp]
\includegraphics[width=0.95\linewidth]{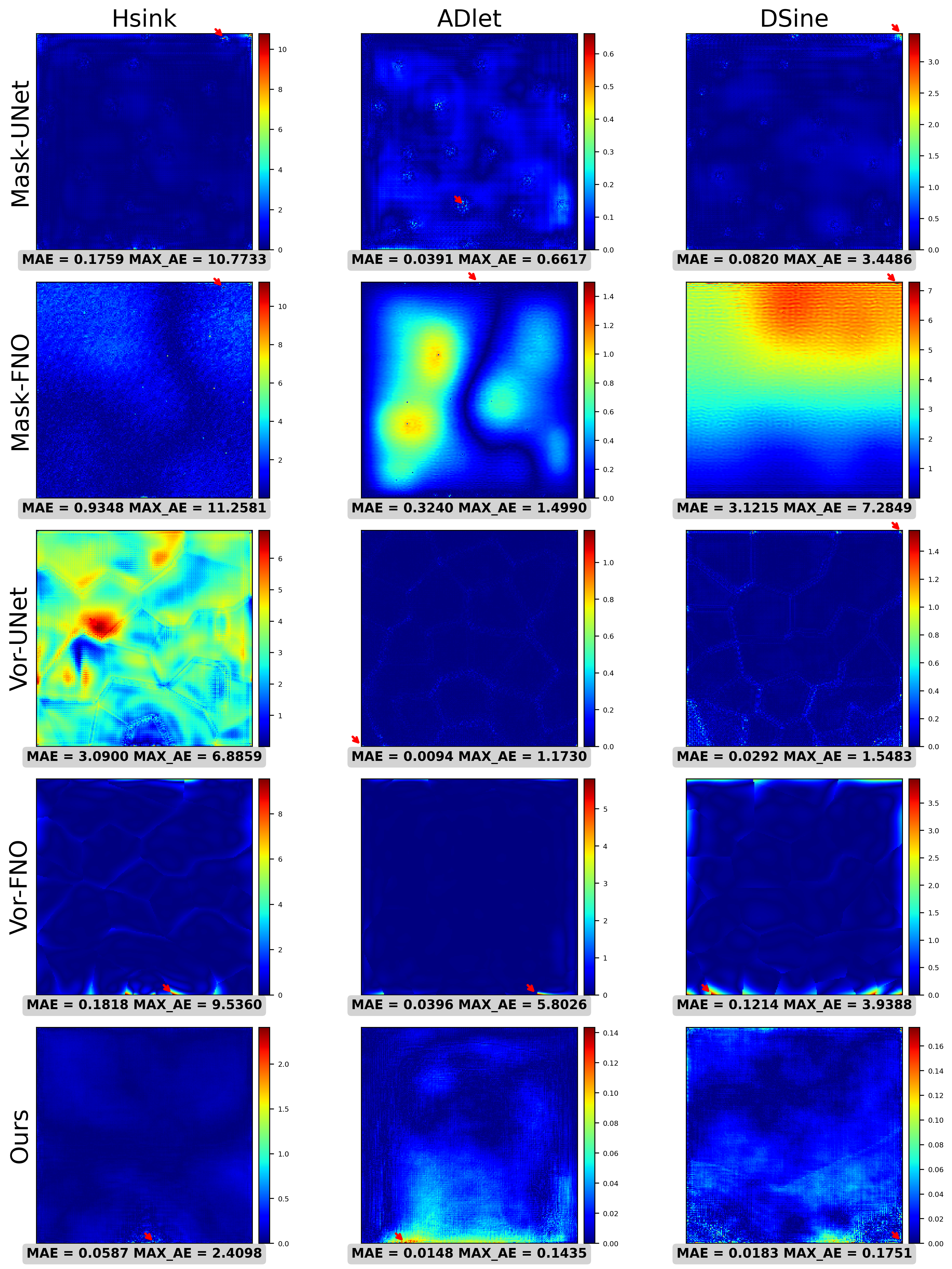}
\caption{The qualitative comparison of our IPTR with other representative methods under single-condition reconstruction.}
\label{fig: all_tfrd_single}
\end{figure}

Qualitative results (Figure \ref{fig: our_tfrd_single} and Figure \ref{fig: all_tfrd_single}) further support these findings. Across all datasets, our reconstructions closely align with the ground truth, with smoother error distributions and fewer high-error zones. This is especially evident in the HSink case, where baseline models such as Mask-FNO and Vor-UNet exhibit visible local artifacts. In ADlet and DSine, our model also preserves finer structural details and maintains uniform error fields. These visual patterns are consistent with the numerical metrics and underscore the reliability of our approach for single-condition reconstruction.

\begin{figure}[htbp]
\includegraphics[width=0.95\linewidth]{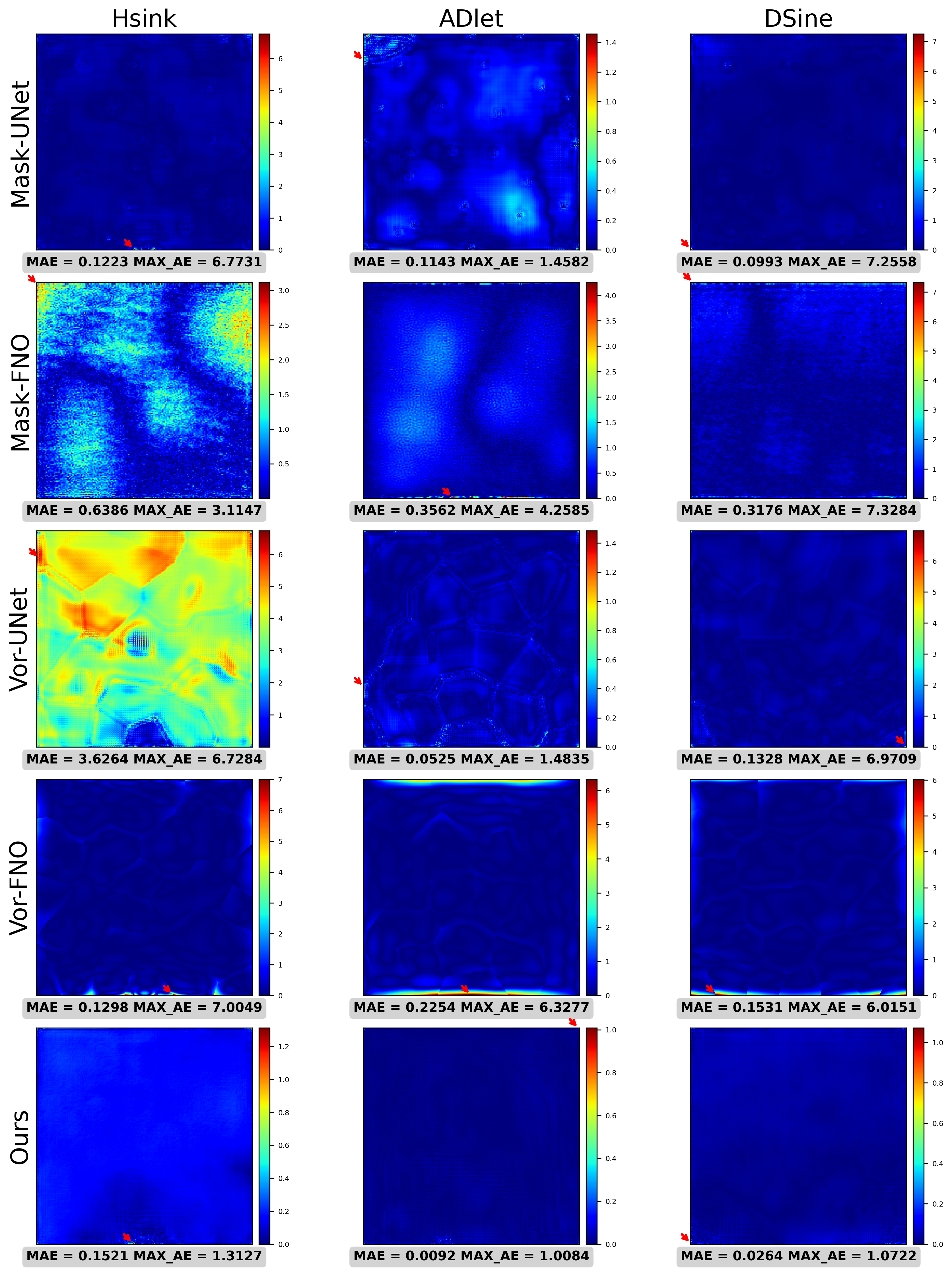}
\caption{The qualitative comparison of our IPTR with other representative methods under multi-condition reconstruction.}
\label{fig: all_tfrd_mix}
\end{figure}

\noindent \textbf{Multi-Condition Reconstruction}: Table~\ref{tab:tfrd-mul} and Figure~\ref{fig: all_tfrd_mix} report the results under the multi-condition setting, where the model is trained on a combined dataset containing samples from all three sub-datasets. Across all test cases, our method consistently outperforms existing approaches in both MAE and Max-AE, achieving the best performance on HSink (0.1375 / 1.7476), ADlet (0.0189 / 1.0168), and DSine (0.0391 / 1.1947). These results highlight the strong generalization ability of our model when exposed to diverse boundary conditions and source patterns during training. The qualitative comparisons in Figure~\ref{fig: all_tfrd_mix} further support these findings. Our model produces spatially smooth and accurate reconstructions, with notably fewer high-error regions. In contrast, other methods—such as Mask-FNO and Vor-UNet—tend to exhibit large localized errors or structural distortions, particularly in the HSink and DSine cases. For instance, Vor-UNet suffers from banding artifacts in HSink and Mask-FNO shows noisy residuals across all scenarios, indicating difficulty in generalizing under mixed-condition inputs. Our results remain stable and visually consistent, demonstrating both high accuracy and reliability in challenging, heterogeneous settings.
\begin{table*}[htbp]
\centering
\resizebox{1\linewidth}{!}{
\begin{tabular}{*{7}{c}} 
\toprule
\multirow{2}{*}{Method} & \multicolumn{2}{c}{\textbf{HSink}} & \multicolumn{2}{c}{\textbf{ADlet}} & \multicolumn{2}{c}{\textbf{DSine}} \\
\cmidrule(lr){2-3} \cmidrule(lr){4-5} \cmidrule(lr){6-7} 
& MAE$\downarrow$ & Max-AE$\downarrow$ & MAE$\downarrow$ & Max-AE$\downarrow$ & MAE$\downarrow$ & Max-AE$\downarrow$ \\
\midrule
Mask-UNet& \underline{0.1559} & 8.4566
 & 0.1026 & 1.5651 & \underline{0.1188} & 9.0847  \\ 
Mask-FNO&0.8158 & \underline{4.6842} & 0.4454 & 3.1466 & 0.4025 & 8.8669  \\ 
Vor-UNet& 6.2315 & 12.9236 & \underline{0.0632} & \underline{1.3899} & 0.1515 & 8.5528  \\ 
Vor-FNO& 0.1860 & 9.5279 & 0.2312 & 6.4576 & 0.1426 & \underline{4.9227}  \\ 
\midrule
Ours & \textbf{0.1375} &  \textbf{1.7476} & \textbf{0.0189} & \textbf{1.0168} & \textbf{0.0391} & \textbf{1.1947} \\
\bottomrule 
\end{tabular}
}
\caption{Quantitative comparisons under multi-condition reconstruction.}
\label{tab:tfrd-mul}
\end{table*}

\subsubsection{Generalization to New Scenario}
\label{newset}
To assess the generalization capability and robustness of IPTR, we construct a new scenario comprising 6000 previously unseen samples that are not included in the TFRD benchmark. Among them, 1000 samples are reserved for validation and another 1000 for testing, while the remaining 4000 samples are used for training with varying data sizes in different experiments. This section presents a series of experiments focusing on sample efficiency and few-shot learning, aiming to simulate real-world deployment conditions where labeled data are limited and sensor layouts are constrained, as exemplified by Case-4 in Figure~\ref{tfr_example}.

\begin{figure}[t]
\includegraphics[width=1\linewidth]{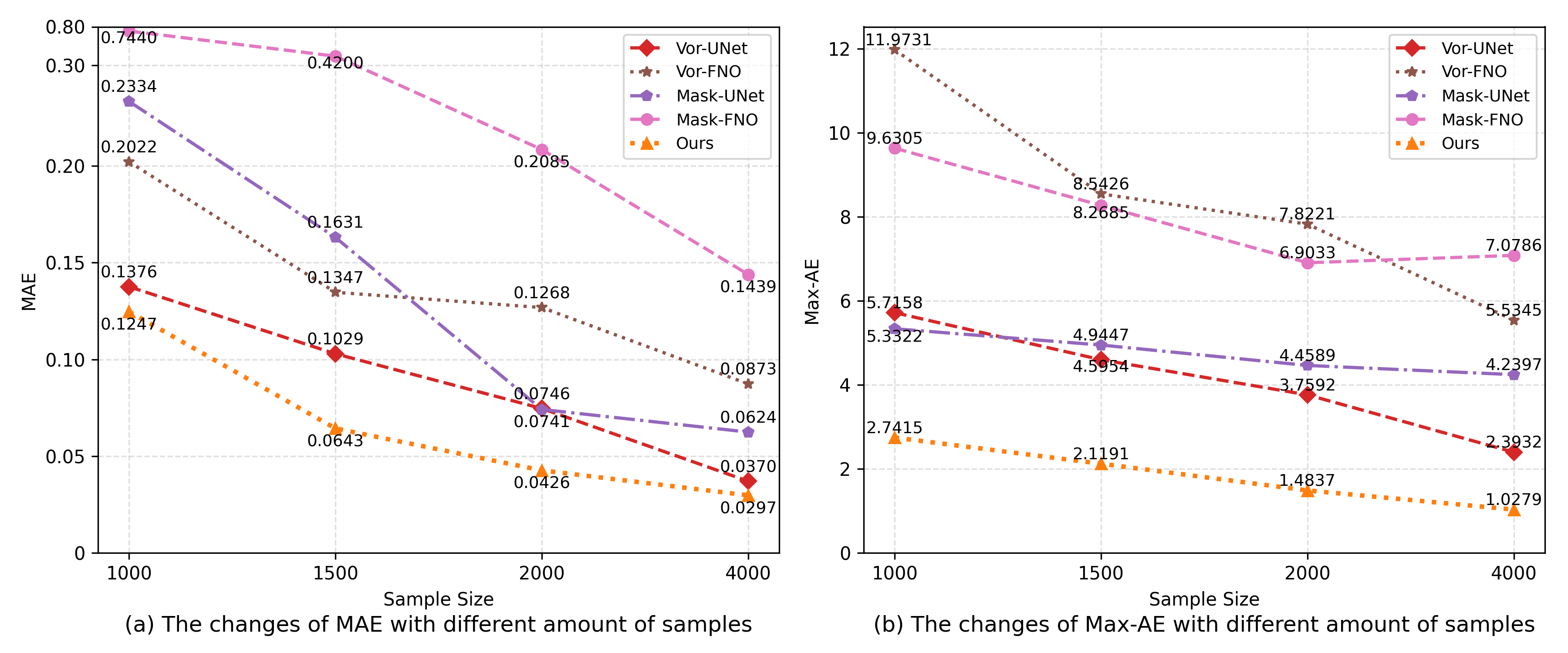}
\caption{Sample-Efficiency Comparison.}
\label{fig: all_new_various}
\end{figure}

\noindent \textbf{Sample-Efficiency Comparison}:
To investigate the sample efficiency of different models under the new scenario, we train each method using varying amounts of labeled data: 1000, 1500, 2000, and 4000 samples. For each training configuration, a corresponding validation set of 100, 200, 500, and 1000 samples is used to monitor performance during training. The quantitative results, visualized in Figure~\ref{fig: all_new_various}, highlight a clear trend: our method consistently achieves the lowest MAE and Max-AE across all sample sizes, demonstrating strong performance across different amounts of samples.

In terms of MAE, our model reaches 0.1247 with 1000 samples and improves to 0.0297 at 4000 samples—surpassing all baselines at every point. Notably, the performance gap widens as the number of training samples increases. For example, when using 4000 samples, the closest competitor (Vor-UNet) reports an MAE of 0.0370, which is still 24.6\% higher than ours. Similarly, Mask-UNet exhibits high variability in performance, starting with poor accuracy at low data volumes and only matching Vor-UNet’s performance at 2000 samples, where our model already achieves superior results. 

When considering Max-AE, our method again leads by a significant margin. With only 1000 samples, our Max-AE is 2.7415—better than almost all other methods even when they are trained with four times more data. At 4000 samples, we achieve a Max-AE of 1.0279, which is less than half that of the best-performing baseline (Vor-UNet: 2.3932), indicating our model's exceptional ability to constrain local errors.

These results collectively suggest that our method not only maintains high accuracy across different training scales, but also scales more effectively with increasing supervision—demonstrating strong generalization under varying amounts of training data. Rather than simulating limited-data conditions, this set of experiments reflects model performance under sufficiently sampled scenarios, allowing us to assess how well different methods benefit from additional supervision. In the next section, we further evaluate the model’s adaptability in few-shot settings by conducting few-shot transfer experiments, where only a limited number of labeled samples are available in the target domain.

\noindent \textbf{Few-Shot Transfer Learning}:
Compared to the previous setting with sufficient training samples, this section evaluates model performance under few-shot condition using transfer learning, which is a practical strategy to leverage prior knowledge when training data in the target scenario are insufficient. All models are initialized from their counterparts pre-trained on the TFRD multi-condition benchmark. This choice reflects a practical consideration: in real-world deployments, the relevance between pre-training and downstream tasks is often unknown or difficult to quantify. Therefore, a model trained across diverse conditions serves as a more general and transferable backbone for downstream adaptation. Fine-tuning is performed with 2, 5, 10, 20, and 50 training samples, each paired with a validation set of 1, 2, 4, 8, and 16 samples, respectively.

\begin{table*}[htbp]
\centering
\resizebox{1\linewidth}{!}{
\begin{tabular}{*{11}{c}} 
\toprule
\multirow{2}{*}{Number of Samples} & \multicolumn{2}{c}{\textbf{2}} & \multicolumn{2}{c}{\textbf{5}} & \multicolumn{2}{c}{\textbf{10}} & \multicolumn{2}{c}{\textbf{20}}& \multicolumn{2}{c}{\textbf{50}}\\
\cmidrule(lr){2-3} \cmidrule(lr){4-5} \cmidrule(lr){6-7} \cmidrule(lr){8-9} \cmidrule(lr){10-11} 
& MAE$\downarrow$ & Max-AE$\downarrow$ & MAE$\downarrow$ & Max-AE$\downarrow$ & MAE$\downarrow$ & Max-AE$\downarrow$ & MAE$\downarrow$ & Max-AE$\downarrow$ & MAE$\downarrow$ & Max-AE$\downarrow$\\
\midrule
Mask-UNet& 3.5720 & 14.6502 & 1.3315 & \underline{13.1989} & 1.2247 & 14.2104 & 0.4999 & 10.4581 & 0.3058 & \underline{7.7115}  \\ 
Mask-FNO &6.5030 & 101.0619 & 2.4550 & 46.4677 & 1.2713 & 12.4416 & 0.7302 & 9.7054 & 0.4021 & 8.8690  \\ 
Vor-UNet& 4.7273 & \underline{13.6441} & 3.8922 & 14.0768 & 3.6637 & \underline{10.1455} & 3.9723 & \underline{8.6465} & 3.3265 & 8.1748  \\ 
Vor-FNO& \underline{0.3892} & 18.4292 & \underline{0.3019} & 16.3998 & 0.2507 & 15.5292 & \underline{0.2012} & 13.0728 & \underline{0.1471} & 9.9537\\ 
\midrule
Ours & \textbf{0.2349} &  \textbf{7.1077} & \textbf{0.1918} & \textbf{7.0079} & \textbf{0.1291} & \textbf{5.7714} & \textbf{0.1049} & \textbf{4.4456} & \textbf{0.0614} & \textbf{3.8818}\\
\bottomrule 
\end{tabular}
}
\caption{Few-shot Transfer Learning.}
\label{tab:few-shot}
\end{table*}

As summarized in Table ~\ref{tab:few-shot}, our model consistently outperforms all baselines across various few-shot scenarios, achieving both lowest reconstruction error and best stability. 

Convolution-based architectures such as Vor-UNet and Mask-UNet exhibit limited transferability in this setting. Vor-UNet, despite integrating geometric priors through Voronoi tessellation, consistently shows high errors across all sample sizes. For instance, with 50 training samples, it still yields a MAE of 3.3265 and a Max-AE of 8.1748. This indicates that its spatial encoding strategy fails to adapt effectively to the new domain. This indicates that its spatial encoding strategy fails to adapt effectively to the new domain. Mask-UNet performs somewhat better as the sample size increases, achieving an MAE of 0.3058 with 50 samples, but it still struggles in the few-shot regime, suggesting that the raw monitoring measurements are insufficient for robust knowledge transfer.

Among operator-based models, Vor-FNO demonstrates relatively stable performance under few-shot transfer. Even with only 10 training samples, it achieves reasonable accuracy, and its errors steadily decrease as the sample size increases. This suggests that its spectral-based encoding retains a certain level of generalization capability. In contrast, Mask-FNO suffers from pronounced instability across all data regimes. Particularly in extremely few-shot settings, such as with only 2 samples, it produces highly unreliable results with a Max-AE exceeding 100, and remains significantly worse than other baselines even at 50 samples. These results indicate that using the original monitoring measurements in Mask-FNO weakens its ability to transfer meaningful representations, likely by disrupting the operator network’s inherent inductive bias.

\begin{figure}[htbp]
\includegraphics[width=1\linewidth]{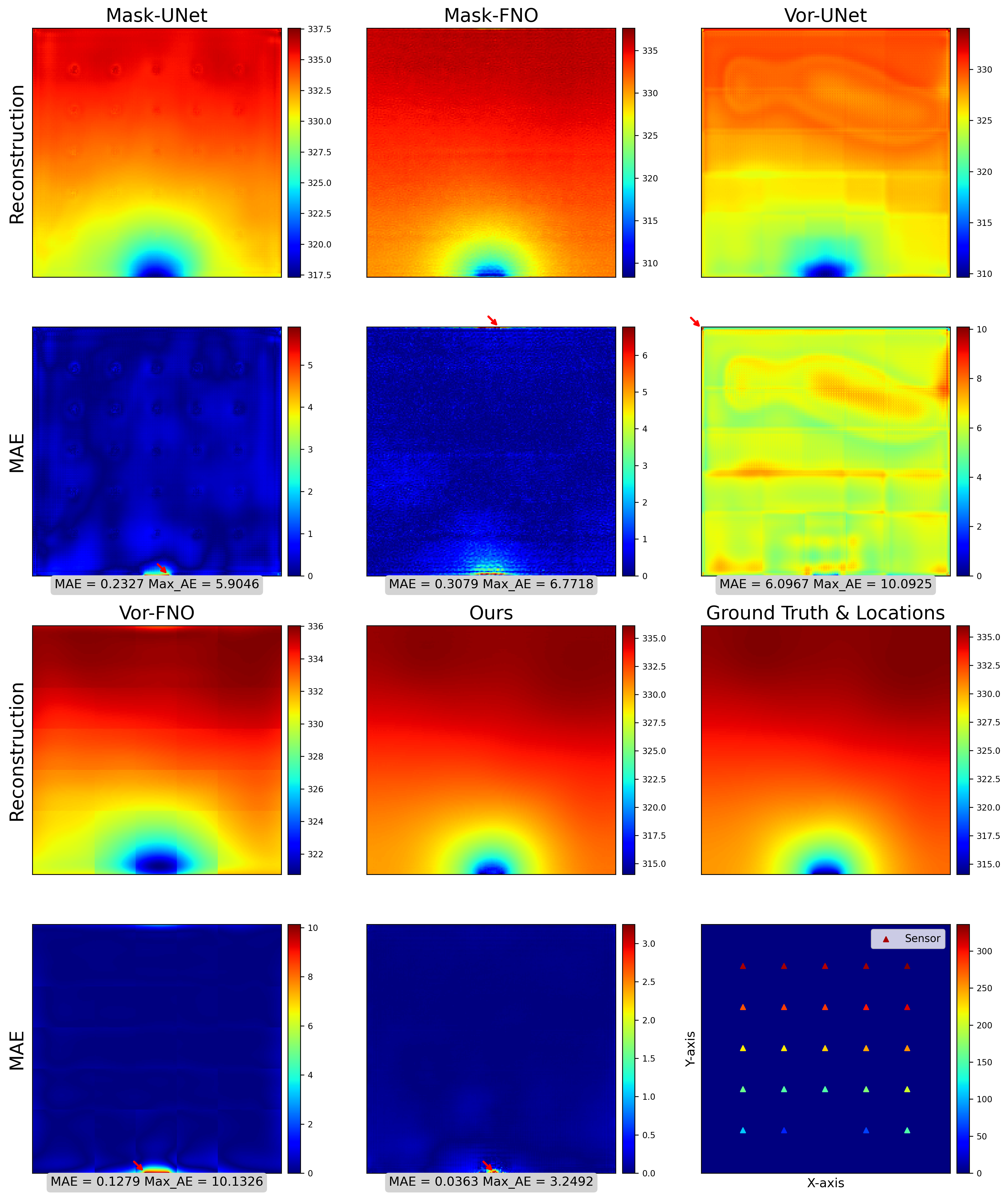}
\caption{Few-shot transfer learning with 50 samples.}
\label{fig: fewshot}
\end{figure}

In contrast, our proposed IPTR consistently outperforms all baseline methods across all few-shot scenarios. With only 2 training samples, IPTR achieves an MAE of 0.2349 and Max-AE of 7.1077. When provided with 50 samples, the performance further improves to an MAE of 0.0614 and Max-AE of 3.8818. More importantly, this performance surpasses many existing methods trained from scratch using substantially larger datasets. For example, Mask-Unet requires 4000 training samples to reach comparable accuracy (MAE of 0.0624), while Vor-FNO with 4000 samples still results in a higher MAE of 0.0873 and Max-AE of 5.5345. To complement the quantitative results, Figure \ref{fig: fewshot} presents the visualization of reconstructed temperature fields and corresponding MAE for all methods when fine-tuned with 50 samples. The visual comparison clearly highlights the superiority of our IPTR framework in capturing detailed temperature variations and preserving structural consistency. In contrast, other baselines show noticeable artifacts, particularly in regions with sharp gradients. This demonstrates that our method significantly enhances data efficiency and transferability.

The superior performance of IPTR is primarily attributed to its architectural design. By incorporating an Implicit Physics-guided branch, IPTR leverages historical simulation data not as direct training input, but as reference to encode implicit physical priors. The dual-branch physics embedding module enables the model to jointly exploit these implicit priors through a cross-attention mechanism, while preserving the target task’s characteristics via a Fourier-based auxiliary encoder. This synergy allows IPTR to maintain physical consistency and stable reconstruction, even when the available training data is extremely limited.
\begin{figure}[htbp]
\includegraphics[width=1\linewidth]{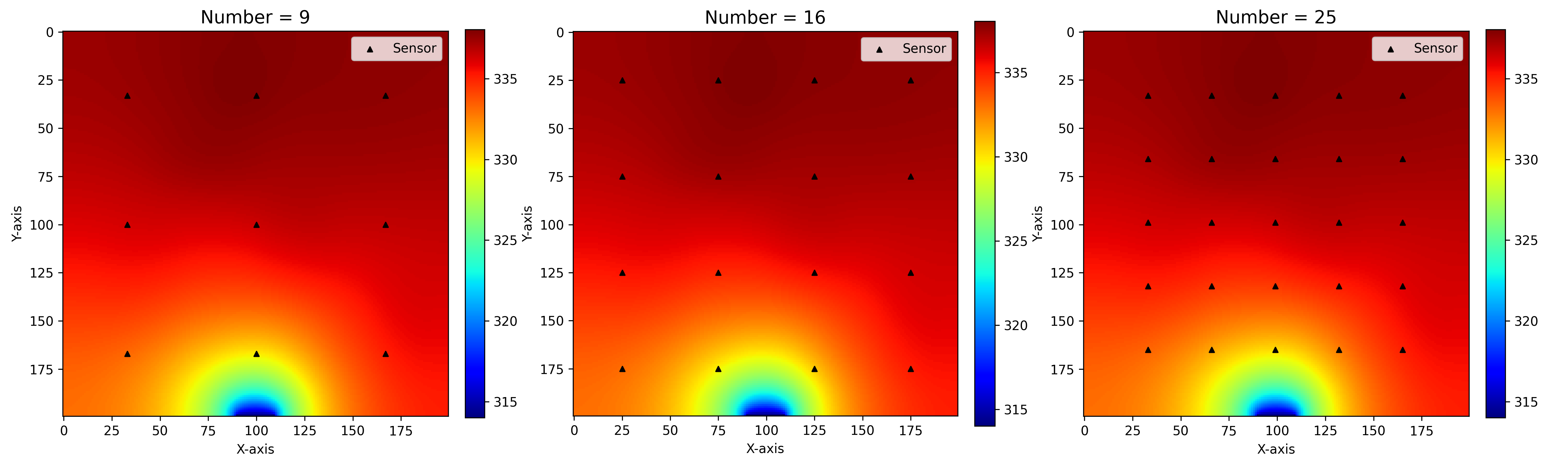}
\caption{Visualizations of different number of sensors.}
\label{fig: num}
\end{figure}
\subsection{Further Discussion}
\label{ablation}
\subsubsection{Influence of the Number of Sensors}
The number of deployed sensors plays a critical role in determining the reconstruction quality of temperature fields, as it directly affects the spatial coverage and informativeness of the input observations. To systematically investigate this factor, we conduct controlled experiments to evaluate the impact of varying sensor quantities on the performance of our proposed method.
All experiments are performed on the newly generated dataset containing 2,000 training samples as illustrated in Section \ref{newset}. To ensure consistency and isolate the effect of sensor quantity, we adopt a uniform placement strategy across all settings. Specifically, we evaluate three sensor configurations (Figure \ref{fig: num}) with 9, 16, and 25 sensors, respectively.
\begin{figure}[htbp]
\includegraphics[width=1\linewidth]{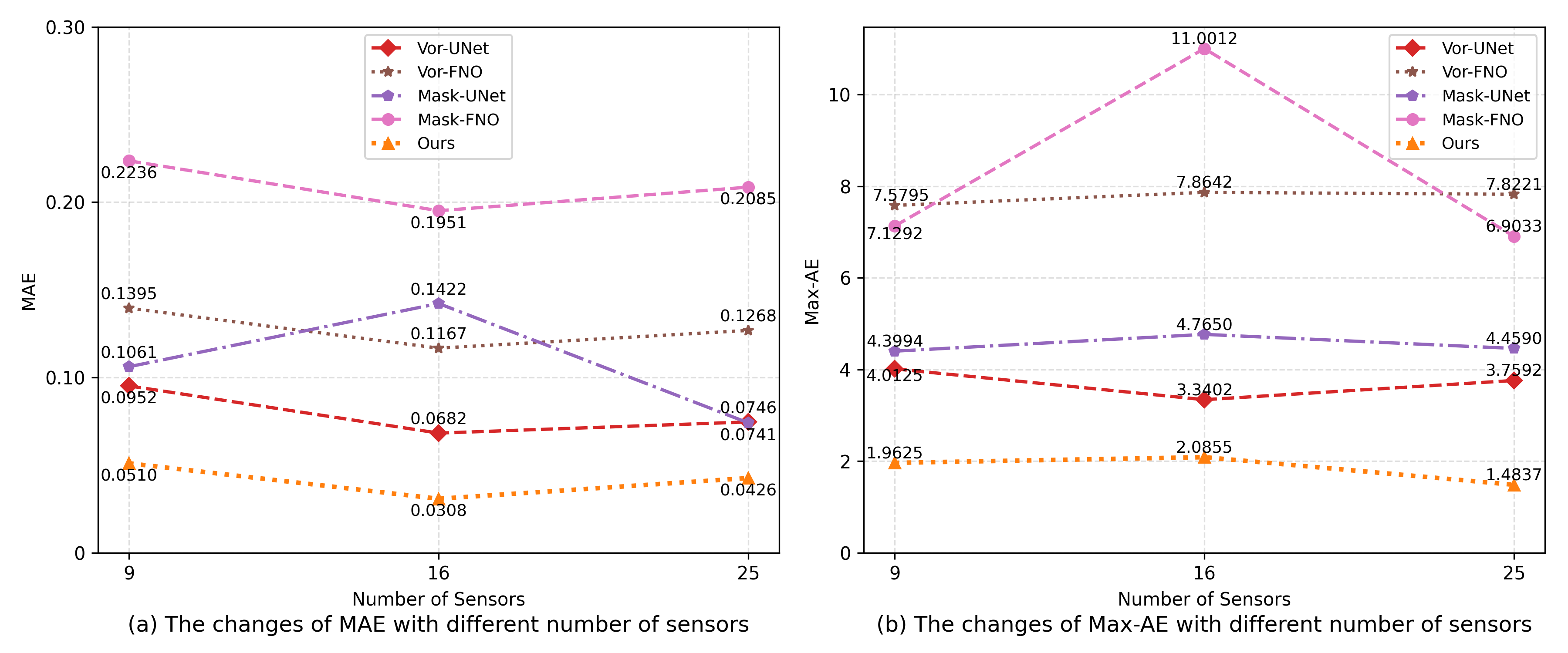}
\caption{Influence of the Number of Sensors.}
\label{fig: sensor_exp}
\end{figure}
The results are summarized in Figure \ref{fig: sensor_exp}. Across all metrics and configurations, our proposed IPTR method consistently outperforms other baselines, achieving the lowest MAE and Max-AE under all sensor counts. Notably, IPTR maintains high reconstruction accuracy even with only 9 sensors, demonstrating its robustness under extremely sparse observation. Furthermore, the reconstruction performance peaks when using 16 sensors, slightly surpassing the 25-sensor case. 
This counter-intuitive observation may be attributed to the potential spatial redundancy introduced by a denser sensor layout. While additional sensors generally provide more information, they may not contribute equally due to overlapping receptive fields or limited additional value in relatively smooth regions of the field. In contrast, the 16-sensor configuration may strike a better balance between spatial coverage and input sparsity, allowing the model to focus on more informative regions without redundancy, thus yielding better generalization.
\subsubsection{Ablation Study}
\label{aba}
To evaluate the effectiveness of our design choices in IPTR, we perform ablation studies on the TFRD multi-condition dataset. Specifically, we assess (1) the impact of different reference sampling strategies for training data construction, and (2) the contribution of the Dual Physics Embedding Module.

\noindent \textbf{The impact of different reference sampling strategy}: Each training pair in IPTR requires a target sample and a corresponding reference sample. We compare two schemes: \textit{Fixed-reference}, where all pairs share a common simulation instance as reference; \textit{Sliding-reference}, where training pairs are formed by sequentially pairing samples with their immediate successors based on dataset index.

Quantitative results as shown in Table \ref{tab:sampling} demonstrate that the sliding-reference sampling strategy significantly outperforms the fixed-reference setting across all sub-tasks. It notably reduces both MAE and Max-AE in HSink and DSine, indicating stronger reconstruction accuracy and robustness. In ADlet, although the Max-AE slightly increases (from 0.2341 to 1.0168), the MAE is substantially reduced (from 0.0457 to 0.0189), suggesting that overall accuracy improves despite rare local deviations. These results confirm that introducing dynamic and diverse reference frames facilitates more effective implicit-physics prior extraction. Accordingly, the sliding-reference strategy is adopted in all subsequent experiments.
\begin{table*}[htbp]
\centering
\resizebox{1\linewidth}{!}{
\begin{tabular}{*{7}{c}} 
\toprule
\multirow{2}{*}{Sampling Strategy} & \multicolumn{2}{c}{\textbf{HSink}} & \multicolumn{2}{c}{\textbf{ADlet}} & \multicolumn{2}{c}{\textbf{DSine}} \\
\cmidrule(lr){2-3} \cmidrule(lr){4-5} \cmidrule(lr){6-7} 
& MAE$\downarrow$ & Max-AE$\downarrow$ & MAE$\downarrow$ & Max-AE$\downarrow$ & MAE$\downarrow$ & Max-AE$\downarrow$ \\
\midrule
Fixed-reference& 0.9820 & 5.4610 & 0.0457 & \textbf{0.2341} & 0.5792 & 1.5822  \\ 
Sliding-reference & \textbf{0.1375} &  \textbf{1.7476} & \textbf{0.0189} & 1.0168 & \textbf{0.0391} & \textbf{1.1947} \\
\bottomrule 
\end{tabular}
}
\caption{Ablation study on sampling strategy.}
\label{tab:sampling}
\end{table*}

\noindent \textbf{Effectiveness of Dual Physics Embedding}:  To investigate the contribution of the proposed Dual Physics Embedding Module, we conducted an ablation study summarized in Table \ref{tab:dpem}. Specifically, we first validate the effect of the implicit physics-guided branch by removing the auxiliary branch (w/o Auxiliary Encoding Branch). This configuration still achieves strong performance across all three sub-tasks, particularly in terms of Max-AE, where the model consistently outperforms all baseline methods in Table \ref{tab:tfrd-mul}. This confirms that the implicit physics priors extracted from reference data provide robust guidance for temperature field reconstruction, even without target spatial encoding. To further assess the role of the auxiliary encoder, we compare the original FNO-based variant with a UNet-based alternative. The model with UNet auxiliary encoder shows degraded performance, especially in HSink and DSine, with significantly increased Max-AE (e.g., 14.44 in HSink), indicating inferior generalization. By contrast, our full model—incorporating the FNO-based auxiliary encoder—achieves consistently strong performance across all tasks and metrics, highlighting the advantage of using Fourier-based representations to encode current field characteristics. 

It is worth noting that removing the implicit physics-guided branch while keeping the auxiliary encoder and identical reference inputs (i.e., w/o Implicit Physics-Guided Branch in Table~\ref{tab:dpem}) does not yield comparable performance. This highlights that the superiority of our method arises not from the use of FNO itself or the naive concatenation of reference data, but from the synergistic design of the dual physics embedding. Among the two branches, the implicit physics-guided branch plays the dominant role by embedding physically consistent priors from reference data, ensuring robust reconstruction even under complex or extreme temperature variations. The auxiliary branch, based on a Fourier backbone, complements this process by refining local details and preserving field-specific characteristics. The dual-branch structure thus provides a balanced yet physics-guided representation that enhances both stability and reconstruction fidelity.

\begin{table*}[htbp]
\centering
\resizebox{1\linewidth}{!}{
\begin{tabular}{*{7}{c}} 
\toprule
\multirow{2}{*}{Method} & \multicolumn{2}{c}{\textbf{HSink}} & \multicolumn{2}{c}{\textbf{ADlet}} & \multicolumn{2}{c}{\textbf{DSine}} \\
\cmidrule(lr){2-3} \cmidrule(lr){4-5} \cmidrule(lr){6-7} 
& MAE$\downarrow$ & Max-AE$\downarrow$ & MAE$\downarrow$ & Max-AE$\downarrow$ & MAE$\downarrow$ & Max-AE$\downarrow$ \\
\midrule
w/o Auxiliary Encoding Branch & \underline{0.3542} & \underline{2.9001} & 0.0547 & \textbf{0.4401} & 0.3475 & 1.4865  \\ 
w/o Implicit Physics-Guided Branch & 0.6641 & 4.2551 & 0.0833 & \underline{0.5611} & 0.2193 & \textbf{1.1333}  \\ 
w/ UNet Auxiliary Encoder & 0.5637 & 14.4374 & \underline{0.0523} & 2.6832 & \underline{0.1228} & 1.7970  \\ 
Ours & \textbf{0.1375} &  \textbf{1.7476} & \textbf{0.0189} & 1.0168 & \textbf{0.0391} & \underline{1.1947} \\
\bottomrule 
\end{tabular}
}
\caption{Ablation study on the role of dual physics embedding module.}
\label{tab:dpem}
\end{table*}

\section{Conclusion}
\label{sec: conclusion}

In this paper, we present IPTR, a novel implicit physics-guided framework for temperature field reconstruction of heat-source systems (TFR-HSS) under sparse sensor observations. Unlike conventional approaches that treat TFR-HSS as a one-to-one regression from sparse monitoring measurements, IPTR introduces a reference monitoring–field pair (under thermal conditions similar to the target) to implicitly encode physical priors. This design enhances the model’s abality to recover extreme regions and improves generalization in data-scarce scenarios.
To effectively integrate target monitoring data and reference information, we design a Dual Physics Embedding Module, consisting of an implicit physics-guided branch and an auxiliary encoder based on Fourier layers. The former extracts latent physical knowledge from reference data via cross-attention, while the latter preserves the spatial structure of target observations. These two branches work in synergy to form a robust representation for accurate field reconstruction.

Extensive experiments are conducted under various conditions—including single-condition, multi-condition, few-shot generalization settings, and different number of sensor placement—demonstrating that IPTR consistently achieves superior reconstruction performance and strong generalization capability in terms of both MAE and Max-AE metrics, validating the effectiveness of implicit physics guidance for solving ill-posed reconstruction problems. In future work, we plan to extend IPTR toward higher-dimensional physical field reconstruction problems, aiming to tackle more complex engineering scenarios with sparse observations. In addition, we will explore more effective strategies for selecting reference pair, beyond the current random sampling scheme, in order to better capture relevant physical correlations and further enhance reconstruction accuracy and robustness.
\section*{Acknowledgments}
This work was partly supported by the Young Elite Scientist Sponsorship Program By CAST (Grant No. YESS20240697) and the National Natural Science Foundation of China (Grant No. 92371206)

{
    \small
    \bibliographystyle{unsrt}
    \bibliography{main}
}

\end{document}